\documentclass{article}

\usepackage{PRIMEarxiv}

\usepackage{bm}
\usepackage{subfigure}
\usepackage{algorithm}
\usepackage{algorithmic}
\usepackage{amsmath}
\usepackage{amsthm}
\newtheorem{definition}{Definition}[section]
\usepackage{amssymb}
\usepackage{hypdoc}
\usepackage{hyperref}       
\usepackage{booktabs}       
\usepackage{graphicx}       
\graphicspath{{./images/}}  

\title{Federated Unlearning via \\ Class-Discriminative Pruning}

\author{
  Junxiao Wang$^\dag$, Song Guo$^\dag$, Xin Xie$^\dag$, Heng Qi$^\ddag$ \\
  Hong Kong Polytechnic University$^\dag$, Dalian University of Technology$^\ddag$ \\
  \texttt{\{junxiao.wang, song.guo, xin-ryan.xie\}@polyu.edu.hk}$^\dag$, \texttt{hengqi@dlut.edu.cn}$^\ddag$ 
}

\begin{document}
\maketitle

\begin{abstract}
We explore the problem of selectively forgetting categories from trained CNN classification models in federated learning (FL).
Given that the data used for training cannot be accessed globally in FL, our insights probe deep into the internal influence of each channel.
Through the visualization of feature maps activated by different channels, we observe that different channels have a varying contribution to different categories in image classification.

Inspired by this, we propose a method for scrubbing the model cleanly of information about particular categories.
The method does not require retraining from scratch, nor global access to the data used for training.
Instead, we introduce the concept of Term Frequency Inverse Document Frequency (TF-IDF) to quantize the class discrimination of channels.
Channels with high TF-IDF scores have more discrimination on the target categories and thus need to be pruned to unlearn. 
The channel pruning is followed by a fine-tuning process to recover the performance of the pruned model.

Evaluated on CIFAR10 dataset, our method accelerates the speed of unlearning by 8.9× for the ResNet model, and 7.9× for the VGG model under no degradation in accuracy, compared to retraining from scratch. 
For CIFAR100 dataset, the speedups are 9.9× and 8.4×, respectively.
We envision this work as a complementary block for FL towards compliance with legal and ethical criteria.
\end{abstract}

\section{Introduction}
Edge devices such as mobile phones and IoT devices have become the primary computing platform today. 
These devices generate a tremendous amount of valuable data while providing hidden insights about the human world. In this case, AI can hugely help evaluate the data and make sense of it. 
By training machine learning models that simulate intelligent behaviours to make well-informed decisions with little or no human intervention. 
However, analyzing large amounts of data with complex machine learning algorithms requires significant computational capabilities. 
Thus, users have to upload their data to a cloud server to train a satisfactory model, which is undesirable from a privacy, security, regulatory or economic perspective.

As an emerging and promising distributed machine learning paradigm, federated learning (FL) allows multiple devices to jointly train a shared model without direct access to sensitive training data \cite{zhao2018federated,kairouz2019advances}. 
In FL, devices train on an initial model locally with their own data, and upload their model updates to the federated server to build a shared global model \cite{mcmahan2017communication,yang2019federated}.
However, in dynamic edge environments, data distribution is often not as static. To allow intelligent applications to adapt to the current environment with limited resources consumption, we need to eliminate specific categories of data from the trained models. 
Moreover, recent privacy legislation such as the European Union's \emph{General Data Protection Regulation} (GDPR) and the former \emph{Right to be Forgotten} \cite{ginart2019making} also give users the right to eliminate specific categories of data from the trained model. 
These practical needs call for efficient techniques that enable FL models to unlearn, or to forget what has been learned from the categories to be removed (referred to as the target categories).

\begin{figure*}[t]
\centering
\includegraphics[width=17.5cm]{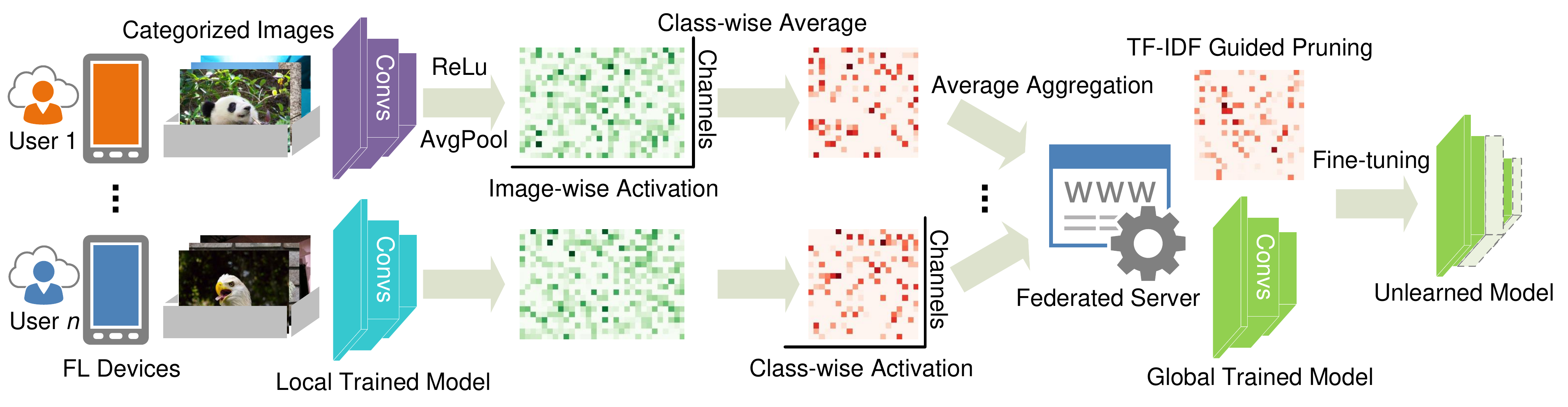}
\caption{Overview and workflow of the proposed unlearning method. Given the GDPR request to remove a specific category, as first, each online FL device downloads an unlearning program from the federated server; Following the program, the local trained CNN model takes the private images as input and generates a feature map score between each channel and category; These feature map scores are then communicated to the federated server and are aggregated as the global feature map scores; The server uses TF-IDF to evaluate the relevant score between the channels and categories, and builds a pruner to execute pruning on the most discriminative channels of the target category; Upon notification of the pruning complete, each online FL device downloads the pruned model from the federated server, and conducts normal federated training program with target category-excluded training data to achieve fine-tuning.}
\label{fig:framework}
\end{figure*}

A naive way to make such models provably forgotten is to retrain them from scratch.
However, the computational and time overhead associated with fully retraining models affected by training data erasure can be prohibitively expensive \cite{bourtoule2021machine}.
Approximate unlearning methods produce ML models that are an approximation of the fully retrained models.
Most of them can be categorized into one of three groups.
Given the category $\bm{i}$ to be unlearned from the trained model.
The \emph{first group} updates the trained ML model using the data $\bm{D_i}$ of $\bm{i}$ to perform a corrective Newton step; it is defined in \cite{guo2020certified}.
The \emph{second group} updates the trained ML model using the remaining data $\bm{D}$-$\bm{D_i}$ to perform a corrective Newton step; it follows \cite{golatkar2020eternal,golatkar2020forgetting,golatkar2021mixed}.
The \emph{third group} updates the trained ML model by correcting the SGD steps that led to the trained model; it follows the method defined in \cite{wu2020deltagrad,wu2020priu}.
Unfortunately, current centralized unlearning methods are not suited to handle the systems challenges that arise in federated learning, including
lack of direct data access, communication cost, and non-IID data. Addressing these challenges is therefore a key contribution of our work.

Through the visualization of feature maps activated by different channels, we observe that different channels have a varying contribution to different categories in image classification.
Inspired by this, we propose a pruning-based method for scrubbing the model cleanly of information about particular categories, as shown in Figure~\ref{fig:framework}. 
The proposed method introduces the Term Frequency Inverse Document Frequency (TF-IDF) to quantize the class discrimination of channels and prune the most relevant channel of the target category to unlearn its contribution to the model.
TF-IDF is a statistical measure that evaluates how relevant a word is to a document in a set of documents. %
Specifically, our unlearning method considers the output of a channel as a word, the feature map of a category as a document, and uses TF-IDF to evaluate the relevant scores between the channels and the categories. 
The channels with high TF-IDF scores have more class discrimination and thus need to be pruned to eliminate the contribution of the target class.
The pruning process does not require any iterative training or searching, and thus is computationally efficient. 
Finally, a fine-tuning process is conducted to recover the performance of the pruned model, and to provide a better utility than the retrained model.

Our approach achieves practical unlearning in FL settings thanks to the following three designs: We primarily consider the \emph{category-level unlearning}, rather than sample-level unlearning. Due to the lack of global access to the raw data, it is impractical for the federated server to evaluate the contribution of each erased data sample. Instead, we will try to evaluate the approximation features aggregated from a small set of participants.
Next, we measure the relationship between channels and categories using a \emph{high-level feature map}, rather than raw data. The high-level feature maps contain more information about the image category. The non-IID raw training data in the same category should also share similar high-level features, thus the class discrimination of the feature map can be learned through collaboration between a small set of participants.
Thirdly, to achieve channel pruning in the federated server, each participant client only needs to upload its \emph{local representation score} that describes the correlation between channels and categories. Such a method avoids raw data leakage and massive communication between client and server, thereby better adapting to the distributed and large scale settings of the FL.

To the best of our knowledge, this is the first time that CNN channel pruning has been used to guide the machine unlearning process. 
We also envision our work as a complementary block for FL towards compliance with legal and ethical criteria.
Extensive experiments on several representative image classification tasks demonstrate the superiority of our unlearning method.
Compared to retraining from scratch, the time required to achieve unlearning is significantly decreased by our method with no degradation in accuracy: on the CIFAR10 dataset, we observe that the speedup is 8.9× for the ResNet model, and 7.9× for the VGG model; on the CIFAR100 dataset, the speedups are 9.9× and 8.4×, respectively.

\begin{figure*}[t]
  \begin{minipage}[t]{1\linewidth}
    \centering
    \includegraphics[height=1.4cm,width=1.8cm]{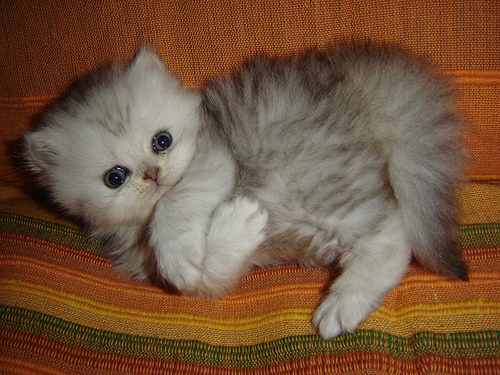}
    \includegraphics[height=1.4cm,width=1.8cm]{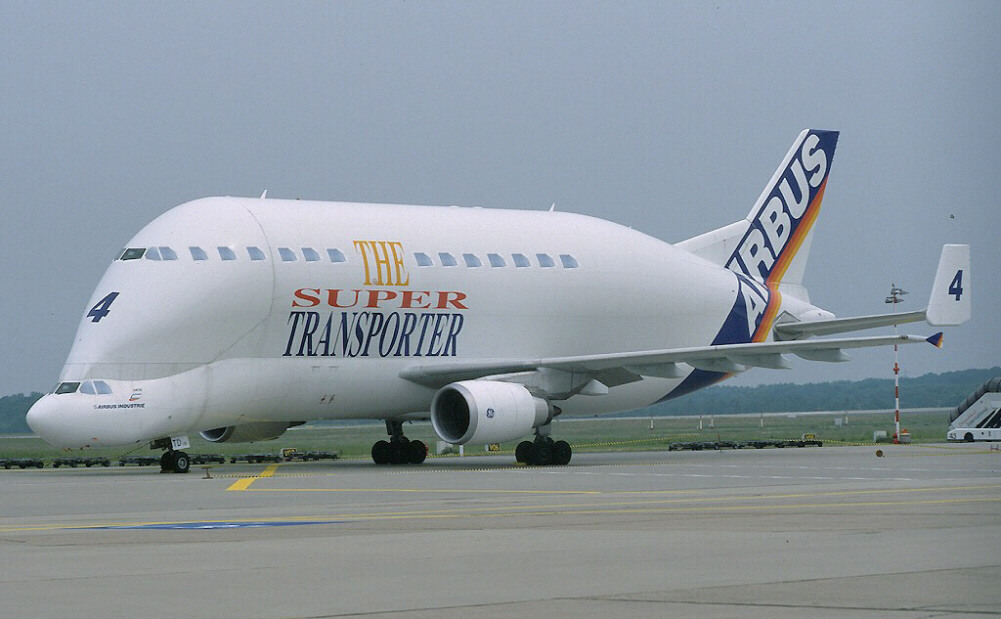}
    \includegraphics[height=1.4cm,width=1.8cm]{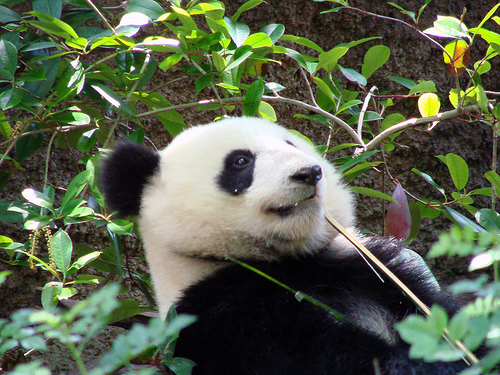}
    \includegraphics[height=1.4cm,width=1.8cm]{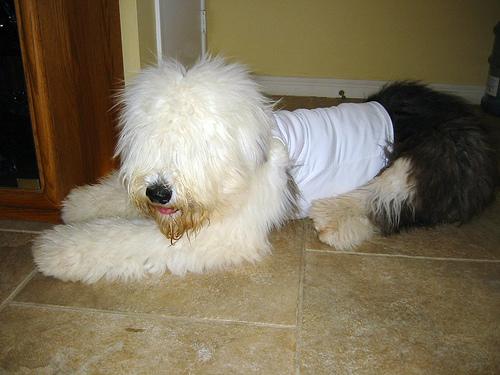}
    \includegraphics[height=1.4cm,width=1.8cm]{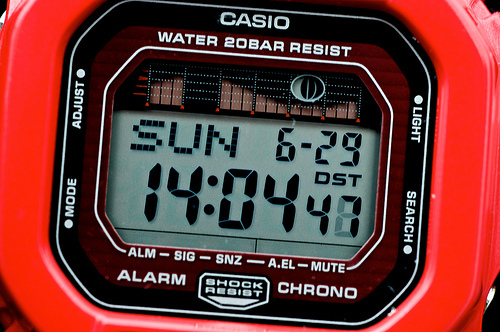}
    \includegraphics[height=1.4cm,width=1.8cm]{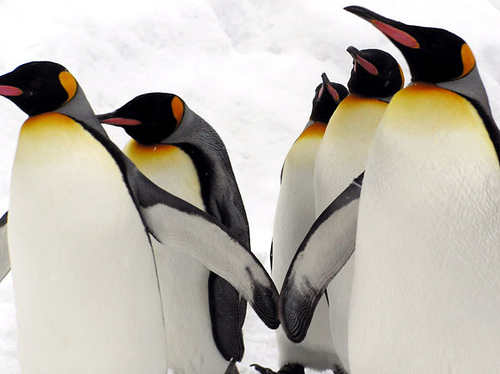}
  \end{minipage}
  \begin{minipage}[t]{1\linewidth}
    \centering
    \includegraphics[height=1.4cm,width=1.8cm]{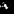}
    \includegraphics[height=1.4cm,width=1.8cm]{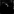}
    \includegraphics[height=1.4cm,width=1.8cm]{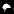}
    \includegraphics[height=1.4cm,width=1.8cm]{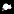}
    \includegraphics[height=1.4cm,width=1.8cm]{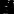}
    \includegraphics[height=1.4cm,width=1.8cm]{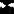}
  \end{minipage}
  \begin{minipage}[t]{1\linewidth}
    \centering
    \includegraphics[height=1.4cm,width=1.8cm]{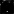}
    \includegraphics[height=1.4cm,width=1.8cm]{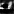}
    \includegraphics[height=1.4cm,width=1.8cm]{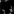}
    \includegraphics[height=1.4cm,width=1.8cm]{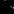}
    \includegraphics[height=1.4cm,width=1.8cm]{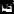}
    \includegraphics[height=1.4cm,width=1.8cm]{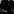}
  \end{minipage}
  \caption{Visualization of images from different categories (first row) along with the feature map of the 52-th and 127-th channels (second and third row, respectively) in the 13-th convolutional layer of VGG16 trained on ImageNet. As seen, with different images as input, head and textual information are always activated by these two channels respectively.}
  \label{fig:visual}
\end{figure*}

\section{Preliminaries}
For sake of completeness, we now briefly review the federated learning (FL), the correspondent problem of target category unlearning, existing studies on unlearning in the context of ML (a.k.a. machine unlearning) and their limitations in the FL settings, due to the inherent distinction in the way how FL and ML learn from data.

\subsection{Federated Learning}
The term \emph{federated learning} was introduced by \cite{mcmahan2017communication}.
Federated learning (FL) is a machine learning setting where a set of $\bm{m}$ clients (\emph{e.g.}, mobile devices) collaboratively train a model under the orchestration of a federated server (\emph{e.g.}, service provider), while the training data of clients is stored locally and not exchanged \cite{kairouz2019advances}. 
The federated server orchestrates the collaborative training process, by repeating the following steps until training is converged:

1) \textbf{Client selection.} Given the unstable client availability, for the round $\bm{t}$ of federated learning, the federated server selects $\bm{n}$ clients meeting eligibility requirements to participate in the learning.

2) \textbf{Local training.} Upon notification of being selected at the round $\bm{r}$, each selected client downloads the current parameters $\bm{w}$ of global model and a training program from the federated server.
Each selected client locally computes an update to the global model on its local training data by executing the training program. 

3) \textbf{Global aggregation.} Upon having received local updates from $\bm{n}$ clients, the federated server aggregates these updates and updates its global model, and initiates next round of learning.

\subsection{General Centralized Unlearning}
\label{subsection:unlearning}
Recently introduced legislation, e.g., the General Data Protection Regulation (GDPR) in the European Union \cite{mantelero2013eu}, introduces a right for individuals to have their data erased \cite{shastri2019seven}.
To remove the $\bm{i}$-th category from a model trained with an integrated dataset $\bm{D}$, a vanilla way would be to retrain the model from scratch on the remaining data points $\bm{D}$-$\bm{D_i}$.
However, the time, computation, and energy costs of model retraining can be quite costly. To address this problem, machine unlearning has been recently studied, which typically produces an approximation of the fully retrained model at low cost. We coarsely characterize these methods into three groups:

1) \textbf{Fisher unlearning method.} The \emph{first group} uses the Fisher information \cite{martens2020new} of the remaining data $\bm{D}$-$\bm{D_i}$ and injects optimal noise in order to unlearn the target category \cite{golatkar2020eternal,golatkar2020forgetting,golatkar2021mixed}. The unlearning algorithm takes as input the currently trained model, the subset of the training data $\bm{D_i}\in\bm{D}$, and outputs an updated model given by
\begin{align}
    \bm{w}'=\bm{w}-\bm{F}^{-1}\bm{\Delta}+\bm{\sigma}\bm{F}^{-\frac{1}{4}}\bm{b}
\end{align} 
where $\bm{\Delta}$ is the gradient of the loss function, computed on the remaining
training data $\bm{D}$-$\bm{D_i}$. 
$\bm{F}$ is the Fisher matrix, also computed on the remaining training data $\bm{D}$-$\bm{D_i}$. 
The first term corresponds to the corrective Newton step. 
The second term corresponds to noise injection, and adds standard normal noise $\bm{b}$ to the updated model in the direction of the Fisher matrix.

2) \textbf{Influence unlearning method.}
The \emph{second group} \cite{guo2020certified,izzo2021approximate} is based on the Influence theory \cite{koh2017understanding}. 
The unlearning algorithm is performed by computing the influence of the data $\bm{D_i}$ on the trained ML model then updating the model to remove that influence as
\begin{align}
    \bm{w}'=\bm{w}+\bm{I}^{-1}\bm{\nabla}
\end{align} 
where $\bm{\nabla}$ is the gradient of the loss function, computed on the data $\bm{D_i}$. 
$\bm{I}$ is the second derivative of the loss function, computed on the remaining training data $\bm{D}$-$\bm{D_i}$. 
The second term is known as the influence function of the data $\bm{D_i}$ on the trained model.
The loss function incorporates noise injection to achieve unlearning.

3) \textbf{Gradient unlearning method.}
The \emph{third group} focuses on approximating the SGD steps as if full retraining was performed \cite{wu2020deltagrad,wu2020priu,graves2021amnesiac,neel2021descent}.
To aid in a history of accurate computations to produce an effective approximation, the unlearning method periodically computes the exact gradient after some iterations. 
The exact gradient is computed on the remaining training data $\bm{D}$-$\bm{D_i}$. 

\subsection{Federated Unlearning}
We primarily consider a scenario where a service provider is required by most users to delete the target data category from their model to protect the user's privacy and avoid legal risks.
We frame the problem of target category unlearning in FL as follows. 
Suppose train a shared model on data points $\bm{D}$ distributed across $\bm{m}$ devices, where $\bm{m}\gg\bm{n}$ (the number of participant devices).
Let $\bm{U}$ $=$ $\{u_1,\cdots,u_i,\cdots,u_{|\bm{U}|}\}$ be the classification space that data space can mapping to, $\bm{w}$ be the model in hypothesis space after pre-training.
To unlearn the $\bm{i}$-th category from the trained model, we need to update the model to make it operate as if the training data $\bm{D_i}$ in $\bm{i}$-th category had never been observed.
\begin{definition}[Federated Unlearning]
We define the algorithm $\mathcal{L}$: $\bm{U}$ $\to$ $\bm{w}$ be a learning process which maps classification space $\bm{U}$ into the hypothesis space of model $\bm{w}$. 
$u'$ be the category that most users require to delete. 
We define an unlearning process $\mathcal{L}^-$: $\mathcal{L}(\bm{U})$ $\otimes$ $\bm{U}$ $\otimes$ $u'$ $\to$ $w'$, which takes an input classification space $\bm{U}$, a learned model $\mathcal{L}(\bm{U})$ and the category $u'$ that required to be forgotten. 
Thus the objective of federated unlearning can be described as
\begin{equation}
    \Phi[\mathcal{L}(\bm{U}\backslash u')] = \Phi[\mathcal{L}^-(\mathcal{L}(\bm{U}), \bm{U}, u')],
\end{equation}
where $\bm{U}\backslash u'$ defines the classification space without target category $u'$. 
$\Phi[\cdot]$ indicates the probability distribution of the output.
\end{definition}
\subsection{Challenges}
The existing centralized unlearning methods always assume that they have access to the relevant global training data (e.g., $\bm{D}$ and $\bm{D_i}$).
However, such assumptions are broken in the FL setting, since the number of participant devices $\bm{n}$ is usually much smaller than the total devices $\bm{m}$ \cite{mcmahan2017communication},
Moreover, the non-IID training data across different participants \cite{zhao2018federated} makes it even more challenging to find the appropriate updating direction for the global models. 
Thus, with incomplete and severely biased local training data, the centralized unlearning method can only offer inaccurate model approximations. 

\section{Pathway to Channel Pruning}
While the data used for training are impossible to access globally in FL, insights may still be gleaned by probing deep into the internal influence of each channel.
Imagine such a possibility: if we can quantify the category information learned by each channel without globally accessing the data, then we have a chance to forget special categories by pruning the channels whose class discrimination is the most. 
Assuming this idea is feasible, the federated unlearning will usher a novel form of implementation, on which the existing practical limitations are well adapted to.
To study the pathway to the channel pruning based solution, we start by investigating the class discrimination of channels in CNNs.

\subsection{Class Discrimination of Channels}
As exploited in \cite{yosinski2015understanding}, the feature map of each channel has the locality that a particular area in one feature map is activated.  
Inspired by this, we visualize the feature map generated by VGG16 trained on ImageNet to explore the local information in the internal 13-th convolutional
layer.
As can be seen from Figure~\ref{fig:visual}, the 52-th channel in the 13-th convolutional layer always generates a feature map that highlights the head information while the 127-th channel always highlights the textual information.
Even though there is no explicitly labeled head or text, the class-discriminative channels automatically learn to extract partial information to make better decisions, which exactly meets human intuition when classifying an image. That is, head information extracted by the 52-th channel helps the model to identify animals, and textual information extracted by the 127-th channel contributes to classifying categories with texts such as airliners and digital watches.
On the other hand, some local features may not be beneficial to identifying all categories. For example, the 127-th channel always chooses to deactivate most of the pixels when processing images with no textual semantics like cats and dogs (see the first and fourth columns).

Such local representation on the internal layer shows that different channels have varying contributions to different categories in image classification, which motivates us to rethink the importance of class-discriminative channels on federated unlearning.
By evaluating the class-discriminative information, we aim to find channels that have the greatest separability with respect to the target categories, which fits seamlessly with the goal of class forgetting.

\subsection{TD-IDF based Channel Scoring}\label{subsec:tf-idf}
We introduce the Term Frequency Inverse Document Frequency (TF-IDF) to quantize the class discrimination of channels.
TF-IDF \cite{paik2013novel} is a statistical measure that evaluates how relevant a word is to a document in a set of documents.
It is used in many scenarios of document search and information retrieval, most importantly in automated text analysis, and is very useful for scoring words in machine learning algorithms for Natural Language Processing (NLP) \cite{yahav2018comments}.
TF-IDF defines a quantification of the relevance between a word and a document: the relevance increases proportionally to the number of times the word appears in the document, but is offset by the number of documents that contain the word. 
Therefore, words that are common in every document, rank low even though they may appear many times, since they mean few to that document in particular.
Instead, if a word appears many times in a document, while not appearing many times in others, it probably means that the word is very relevant.

Concretely, TF-IDF for a word in a document is quantized by multiplying two different metrics: (1) The term frequency (TF) of the word in the document; (2) The inverse document frequency (IDF) of the word across a collection of documents. 
For word $\bm{t}$ in document $\bm{e}$ from the set $\bm{E}$ of all documents, TF is usually calculated with a count of instances of $\bm{t}$ appearing in $\bm{e}$. Then, the frequency is divided by the length of $\bm{e}$.
The IDF can be calculated by taking the total volume of $\bm{E}$, dividing it by the number of documents that contain $\bm{t}$, and calculating the logarithm.
Multiplying the TF and IDF results in the TF-IDF score of $\bm{t}$ in $\bm{e}$. 

Determining how discriminative a channel is to a category can use TF-IDF in a similar way.
Specifically, our unlearning method regards the output of a channel as a word and the feature map of a category as a document.
Hence, this TF-IDF variant can evaluate the relevant score between the channels and the categories.
Channels with high TF-IDF scores have more class discrimination and thus need to be pruned. 

\subsection{Class-Discriminative Channel Pruning}
Channel pruning targets at removing the specific channels in each layer and shrinking a CNN model into thinner one \cite{he2017channel}.
As a type of structured simplification, the advantage of channel-level pruning is removing the entire channels directly, which can be well supported by general-purpose hardware and high-efficiency Basic Linear Algebra Subprograms (BLAS) libraries \cite{lin2020channel}.

Moreover, existing methods of class-discriminative channel pruning have not well studied the computational effectiveness.
For example, the recent work \cite{kung2019methodical} uses a closed-form function and the work \cite{zhuang2018discrimination} inserts a cross-entropy type loss.
They both require iterative optimization steps or heavy matrix operation like matrix inversion to obtain the discrimination score, which can hinder these methods to be applied to large-scale FL settings.
Compared to existing methods of class-discriminative channel pruning, our TF-IDF guided channel pruning does not require any iterative training or searching, which is much more lightweight in terms of computational overhead.

\section{Federated Unlearning Framework}\label{subsec:framework}
Our framework is applicable to FL models that pre-trained in a wide range of ways, unlike existing unlearning methods that require the training to be conducted in a special manner. 
Given the request to remove a specific category from a pre-trained model stored in both local and global, the workflow depicted in Figure~\ref{fig:framework} is composed of two key components running respectively in the FL client and the federated server: participant clients transform their private images to generate local representations which are communicated to the server, in order to build a pruner based on the TF-IDF score between the target category and channels.

\subsection{Local Processing in FL Clients}
\label{subsec:local_proc}
At each FL client, \texttt{local\_proc} in the unlearning program is conducted with local private images, in order to generate a local representation between channels and classes.

Considering the local trained model as an $L$-layer CNN model, whose kernel weights can be represented as $\bm{w}$ = $\left\{\bm{w}_1, \bm{w}_2, \dots, \bm{w}_L\right\}$. 
The kernel in the $l$-th layer is denoted as $\bm{w}_l\in\mathcal{R}^{C_{out}^l\times C_{in}^l\times K_l \times K_l}$, where $C_{out}^l$, $C_{in}^l$, $K_l$ denote the numbers of output channels and input channels, and the kernel size, respectively.
Let $\bm{X}_l\in\mathcal{R}^{N\times C_{in}^l\times H_l \times W_l}$ be the input of the $l$-th layer where $N$ is the batch size of input images, and $H_l$, $W_l$ are the height and width of the input respectively.
Therefore, the local set of private images can be denoted as $\bm{X}_1$.
The output feature map of the $l$-th layer is calculated as
\begin{align}\label{equ:feature}
    \bm{O}_l=\bm{X}_l\circledast \bm{w}_l
\end{align} 
where $\circledast$ is the convolutional operation, $\bm{O}_l\in\mathcal{R}^{N\times C_{in}^{l+1}\times H_{l+1} \times W_{l+1}}$, and $C_{in}^{l+1}$=$C_{out}^l$.
As a first step, the client will record the feature map generated by local model in each layer.
Then, we apply ReLU \cite{nair2010rectified} followed by an average pooling operation over the feature map. 
For the channels at $l$-th layer, the activation of output feature map is denoted by $\bm{A}_l$, can be calculated by
\begin{align}
    \bm{A}_l=\text{AvgPooling}(\text{ReLU}(\bm{O}_l))
\end{align} 
where $\bm{O}_l$ is the output feature map generated by $l$-th layer, and the feature map of each channel is resized from $H_{l+1}$$\times$$W_{l+1}$ to 1$\times$1. 
Therefore, $\bm{A}_l\in\mathcal{R}^{N\times C_{out}^{l}}$ is a local representation between images and channels. 
Finally, the representation $\bm{A}_l$ will be averaged across classes and re-stacking each channel, as $\bm{A}'_l\in\mathcal{R}^{|\bm{U}|\times C_{out}^{l}}$, where $|\bm{U}|$ is the number of classes.
Upon the unlearning program is conducted completely in the client, the representation $\bm{A}'_l$ is uploaded to the federated server for further processing.

\subsection{Processing in the Federated Server}
\label{subsec:server_proc}
At the federated server, \texttt{server\_proc} is conducted with the local uploaded representations between classes and channels.
Upon having received local representations from all participant clients, the server first aggregates these local representations in average to generate a global representation between classes and channels.

Given a global representation $\bm{A}^*\in\mathcal{R}^{|\bm{U}|\times C_{out}^{l}}$ in $l$-th layer, the term frequency (TF) can be calculated by
\begin{align}\label{equ:tf}
    \text{TF}_l^{u'}=\frac{{\bm{A}^*}_l^{u'}}{{\bm{\sum}}_{j=0}^{C_{out}^l}{\bm{A}^*}_l^{u',j}}
\end{align}
where TF$_l^{u'}\in\mathcal{R}^{C_{out}^l}$ represents the contribution of each word (channel) to a specific document (class $u'$).
Note that, some channels that have high scores in TF may also have a contribution to other categories outside the target category. 
In order to obtain the most discriminative channels of the target category, the inverse document frequency (IDF) in $l$-th layer can be calculated by   
\begin{align}\label{equ:idf}
    \text{IDF}_l^j=\log\frac{1+|\bm{U}|}{1+\left|\big\{u_i\in\bm{U}: {\bm{A}^*}_l^{u_i,j}\geq\text{Avg}\big({\bm{A}^*}_l^{u_i}\big)\big\}\right|}
\end{align}
where IDF$_l^j\in\mathcal{R}^1$ represents how common or rare contribution of a specific word (channel $j$) is in the entire document set (classes).
The closer it is to 0, the more common contribution of a channel is.
Hence, if contribution of a channel is very common and appears in many classes, its IDF will approach 0. Otherwise, it will approach 1.
Multiplying TF$_l^{u'}$ and IDF$_l$ results in the TF-IDF$_l^{u'}$ score of channels for the target class in $l$-th layer, as
\begin{align}\label{equ:tf-idf}
    \text{TF-IDF}_l^{u'}=\text{TF}_l^{u'}*\{\text{IDF}_l^1,\cdots,\text{IDF}_l^{C_{out}^l}\}
\end{align}
where TF-IDF$_l^{u'}\in\mathcal{R}^{C_{out}^l}$ represents the relevant score between the channels and categories.
The higher the score, the more relevant that channel is in that particular class.

Based on the calculated TF-IDF score between channels and categories, the server then builds a pruner to execute pruning on the most discriminative channels of the target category. 
Pruning is a common technique to compress neural network models. 
It prunes specific weights from models, thereby their values are zeroed and we make sure they do not take part in the back-propagation process.
We adopt a one-shot pruning to prune the channels whose TF-IDF score is beyond a pre-defined percentage $R$. 
Since some channels in $l$+1-th layer are removed, the convolution filters that generate these channels in $l$-th layer will be removed accordingly. 
The pruned kernel $\hat{w}_l\in\mathcal{R}^{\hat{C}_{out}^l\times \hat{C}_{in}^l\times K_l \times K_l}$ is obtained under the constraints of $\hat{C}_{in}^l$$\leqslant$$C_{in}^l$ and $\hat{C}_{out}^l$$\leqslant$$C_{out}^l$. 
We have the Equation~\ref{equ:feature} in the pruned model $\hat{\bm{w}}_l$ as
\begin{align}
    \hat{\bm{O}}_l=\hat{\bm{X}}_l\circledast \hat{\bm{w}}_l
\end{align} 

In the case of multi-class removal, the pruning process is executed multiple times, removing one class each time. 
Finally, upon the pruning complete, the federated server will notify each participant FL client to download the pruned model from it.
A fine-tuning process is then conducted to achieve the after-pruned model with a target accuracy.

\subsection{Fine-tuning Processing}
\label{subsec:finetuning_proc}
After pruning the most discriminative channels of the target category, the accuracy degradation should be compensated by retraining the pruned model.
Our fine-tuning process is the same as the normal training procedure of federated learning, without introducing additional regularization.
To reduce the unlearning time, we apply the \emph{prune once and retrain} strategy: prune channels of multiple layers at once and retrain them until the target accuracy is restored.  
We find for our unlearning method, the \emph{prune once and retrain} strategy can be used to prune away significant portions of the model corresponding to the target class, and any loss in accuracy can be regained by retraining for a short period of time (significantly less than retraining from scratch). 

\subsection{Discussion}
\textbf{Feasibility and limitations of applying our method to centralized learning}.
Although specifically designed for FL, the class-discriminative channel pruning method can definitely be applied in centralized learning, since the representative feature distribution between channels and categories can easily be obtained with global access to the data.
However, the advantages of class-discriminative pruning are significantly diminished in centralized scenarios. With full access to the raw data, the central server can accurately evaluate the individual contributions of each data point. Additionally, data privacy protection and communication overhead optimization is no longer required. 
This will lead to a far greater diversity of class-unlearning designs, allowing the central server to obtain a better approximation to the target unlearned model in terms of accuracy or overhead.

\textbf{Challenges on sample-level unlearning}.
This study focuses on class-level unlearning problems in FL scenarios, while the sample-level unlearning problem is a more challenging issue that remains to be studied in future work.
The sample-level unlearning task requires the model to remove specific data samples from the model while still maintaining the accuracy of the model.
To achieve such an ambitious goal, it requires a more elaborate design since the data point contributions to the model are difficult to evaluate without access to the raw data, and we are working on a solution to the problem in the near future.

\section{Experimental Evaluation}
In this section, we empirically evaluate the performance of the proposed unlearning method in the FL settings. 
We report experiments on CIFAR10 and CIFAR100 datasets \cite{krizhevsky2009learning} with ResNet and \cite{he2016deep} VGG \cite{simonyan2014very} models.
We choose these datasets and models because they are sufficiently standard and general in image classification tasks, thus proving that our method is general instead of only valid for specific image tasks, such as facial image classification. 
Specifically, CIFAR10 dataset consists of 60000 32x32 colour images in 10 classes, with 6000 images per class.
CIFAR100 dataset is just like the CIFAR10, except it has 100 classes containing 600 images each.

\subsection{FL Settings}
Note that, the characteristic of FL is that only a portion of total clients are sampled as participant, and the participant clients have Non-IID local training data \cite{mcmahan2017communication,konevcny2016federated,zhao2018federated}.
Therefore, the participant data are incomplete and biased.
We assume the participant data have a distribution different from the overall training data distribution of the learning task.
In particular, we assume the participant data are biased towards certain classes.

\begin{figure*}[t]
  \begin{minipage}[t]{1\linewidth}
    \centering
    \includegraphics[width=0.23\columnwidth]{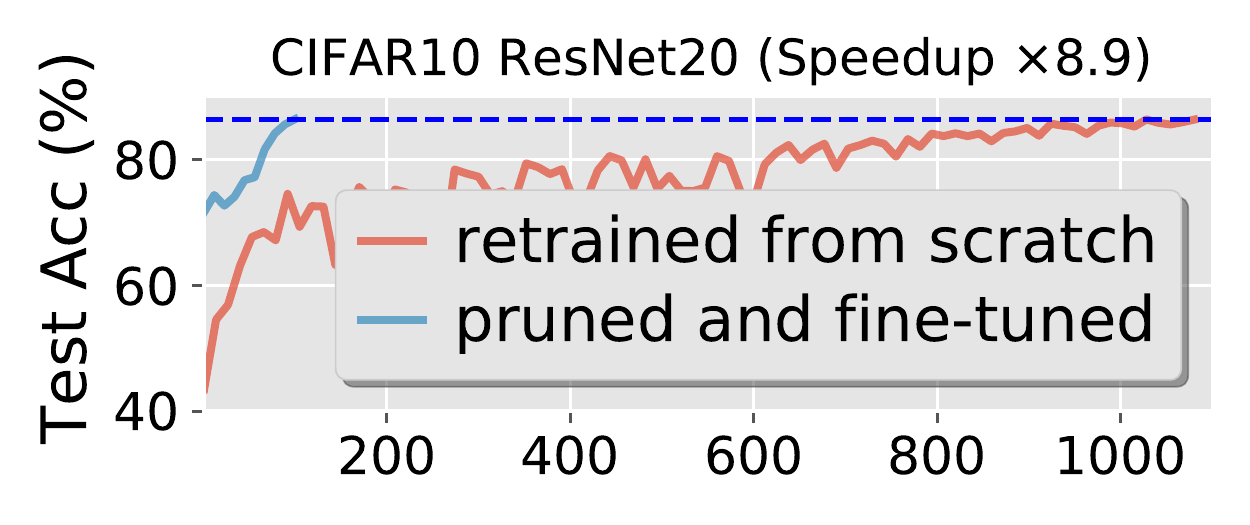}
    \includegraphics[width=0.23\columnwidth]{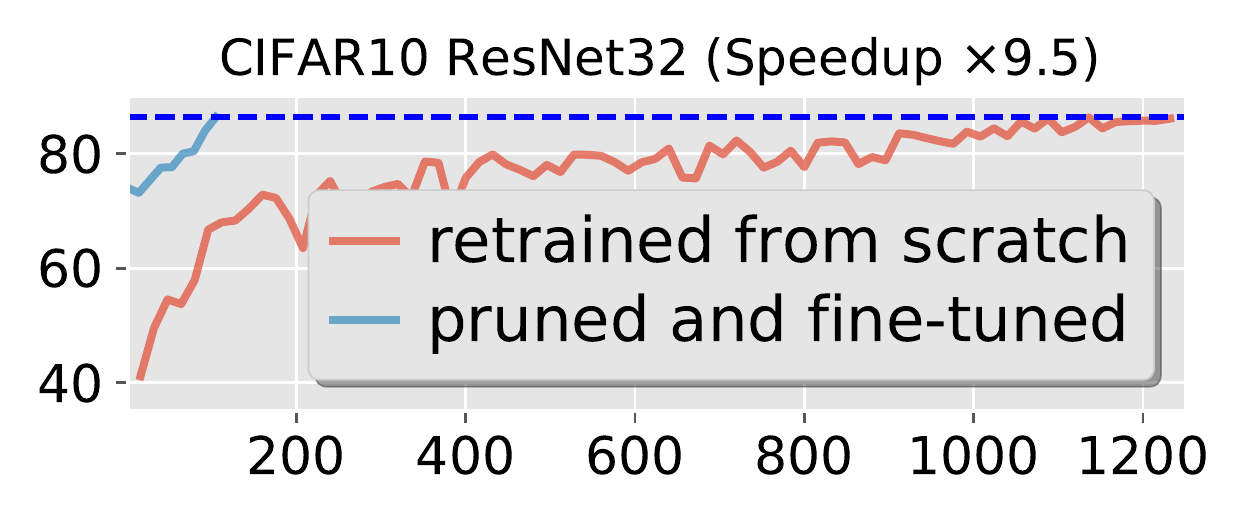}
    \includegraphics[width=0.23\columnwidth]{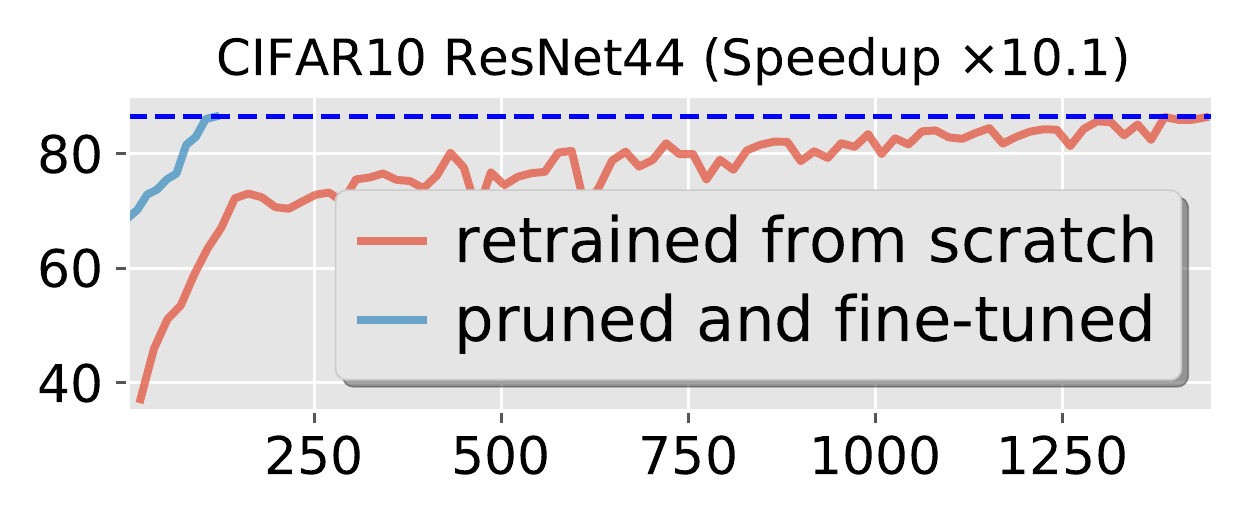}
    \includegraphics[width=0.23\columnwidth]{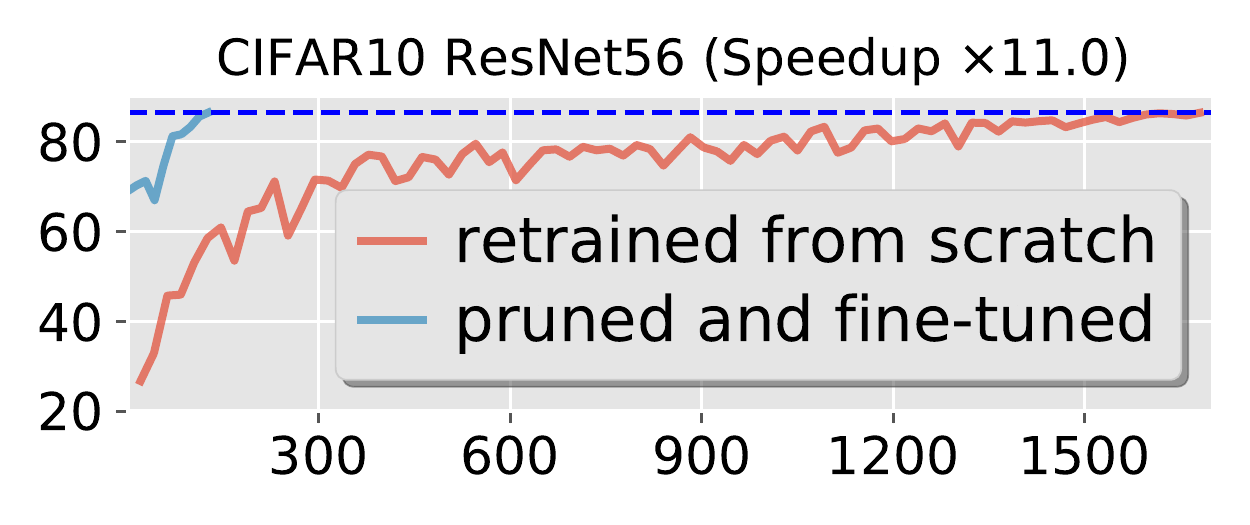}
  \end{minipage}
  \begin{minipage}[t]{1\linewidth}
    \centering
    \includegraphics[width=0.23\columnwidth]{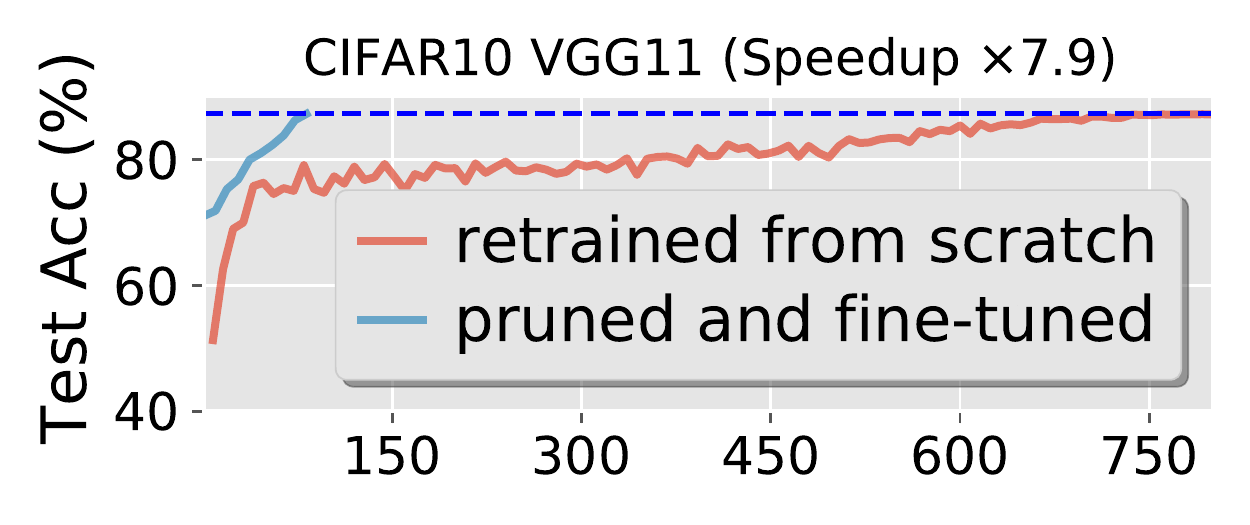}
    \includegraphics[width=0.23\columnwidth]{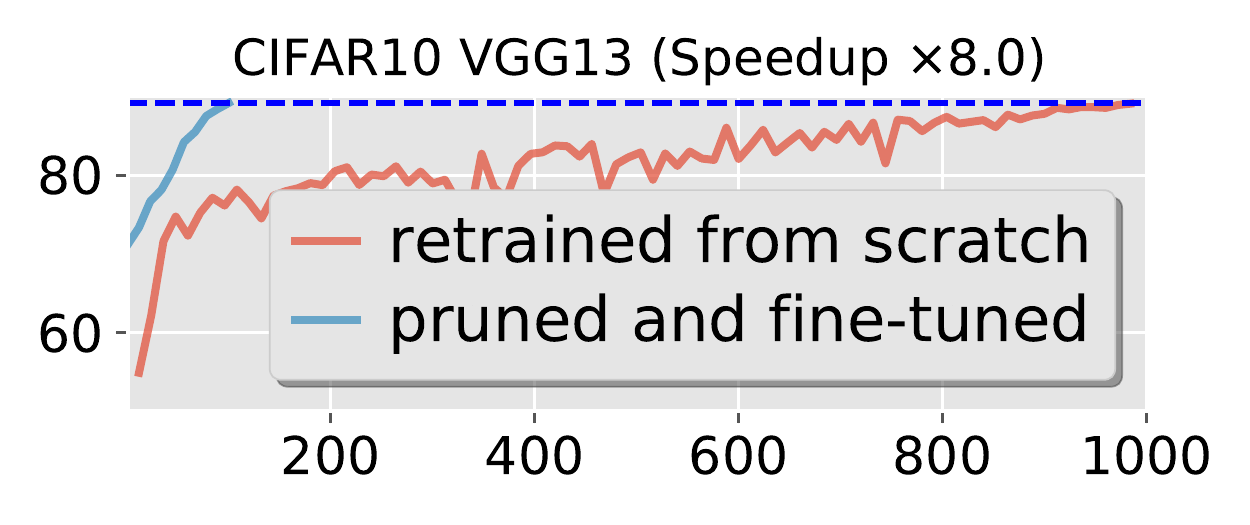}
    \includegraphics[width=0.23\columnwidth]{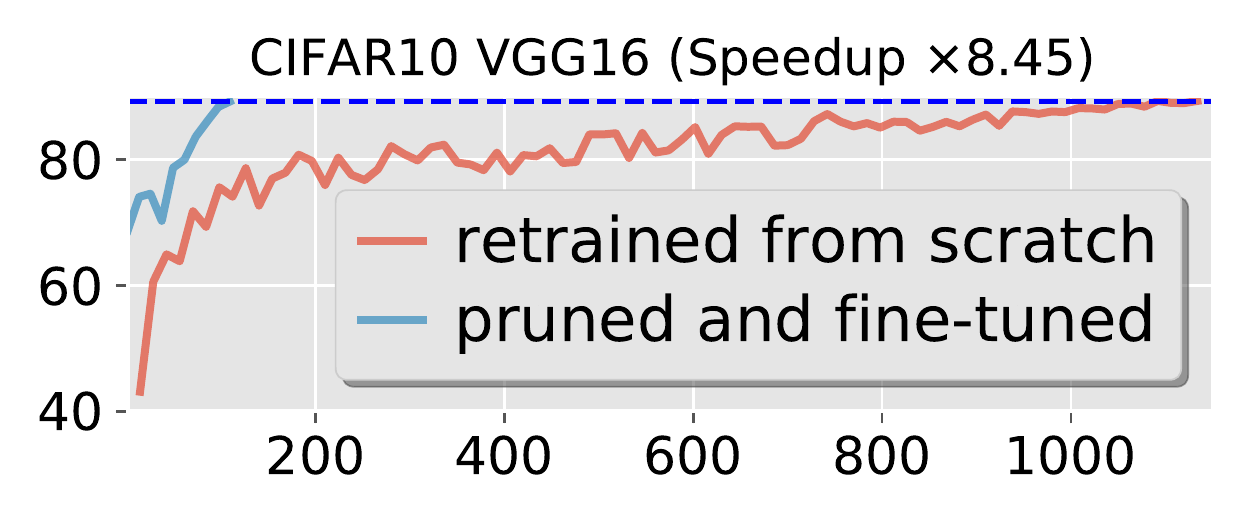}
    \includegraphics[width=0.23\columnwidth]{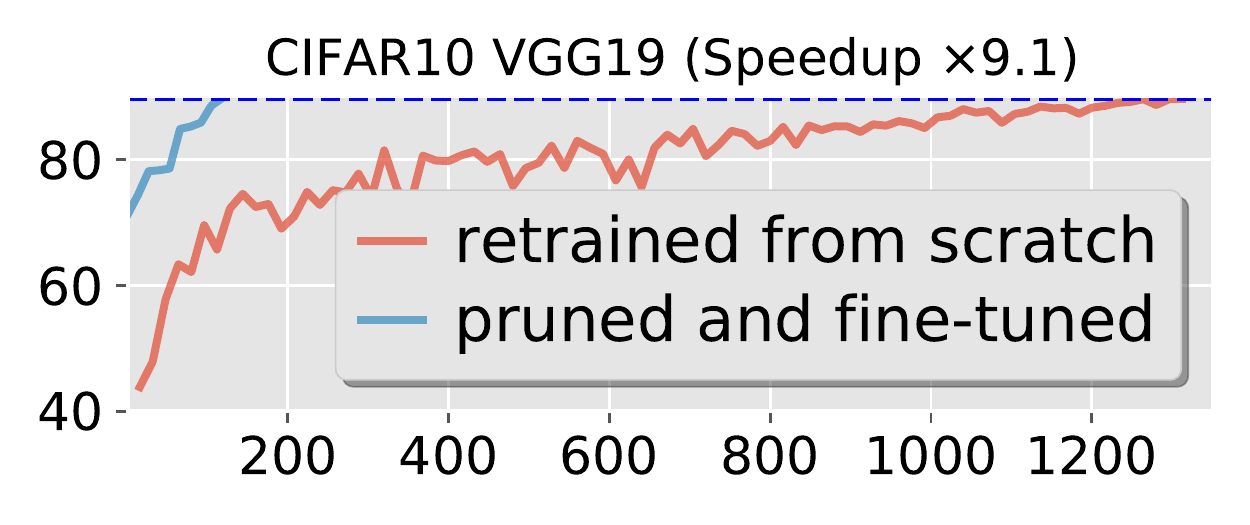}
  \end{minipage}
  \begin{minipage}[t]{1\linewidth}
    \centering
    \includegraphics[width=0.23\columnwidth]{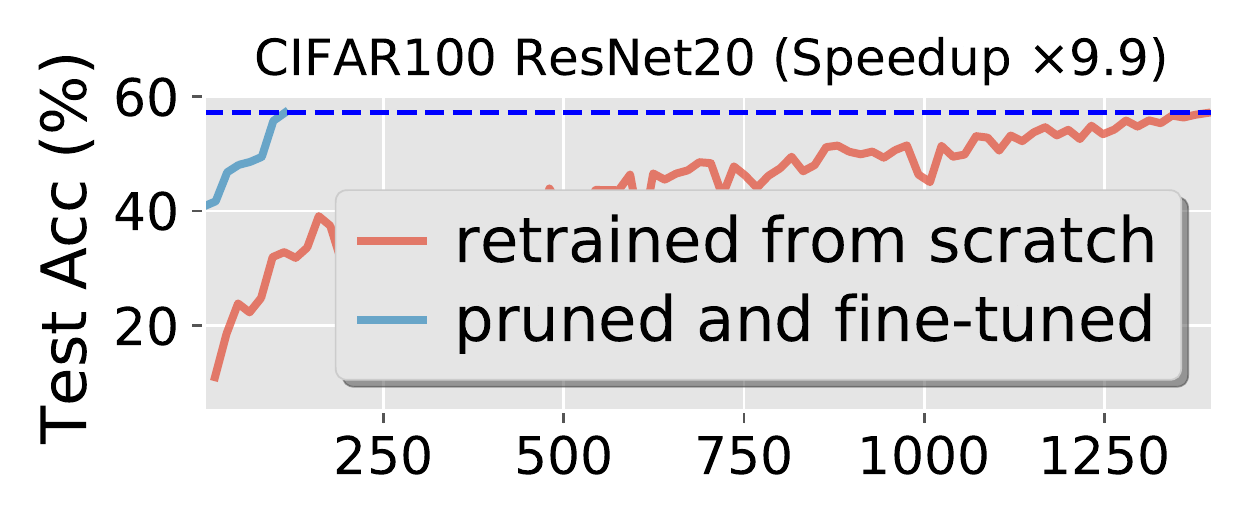}
    \includegraphics[width=0.23\columnwidth]{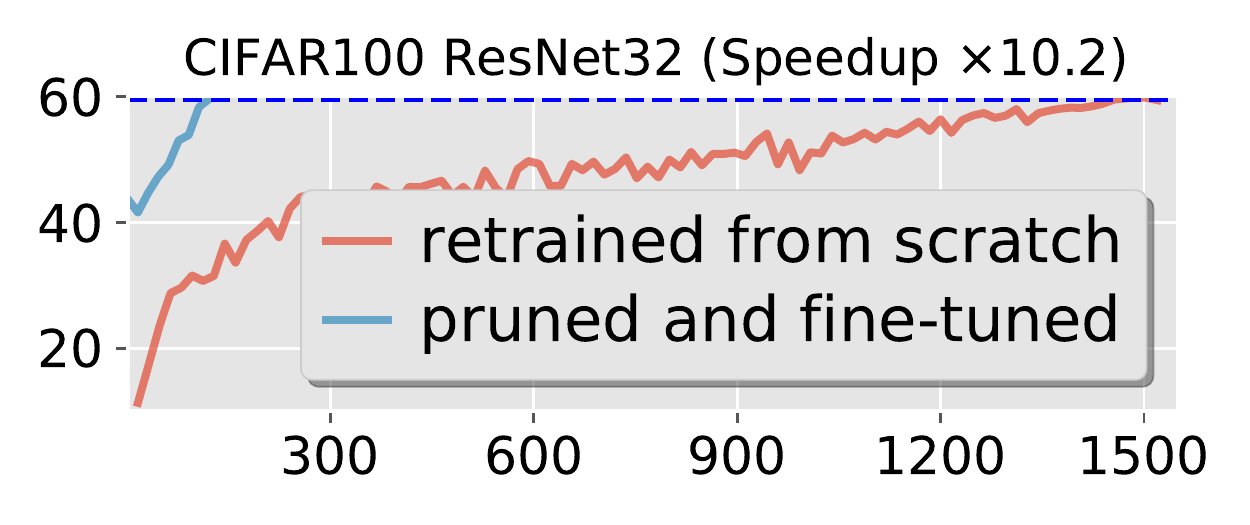}
    \includegraphics[width=0.23\columnwidth]{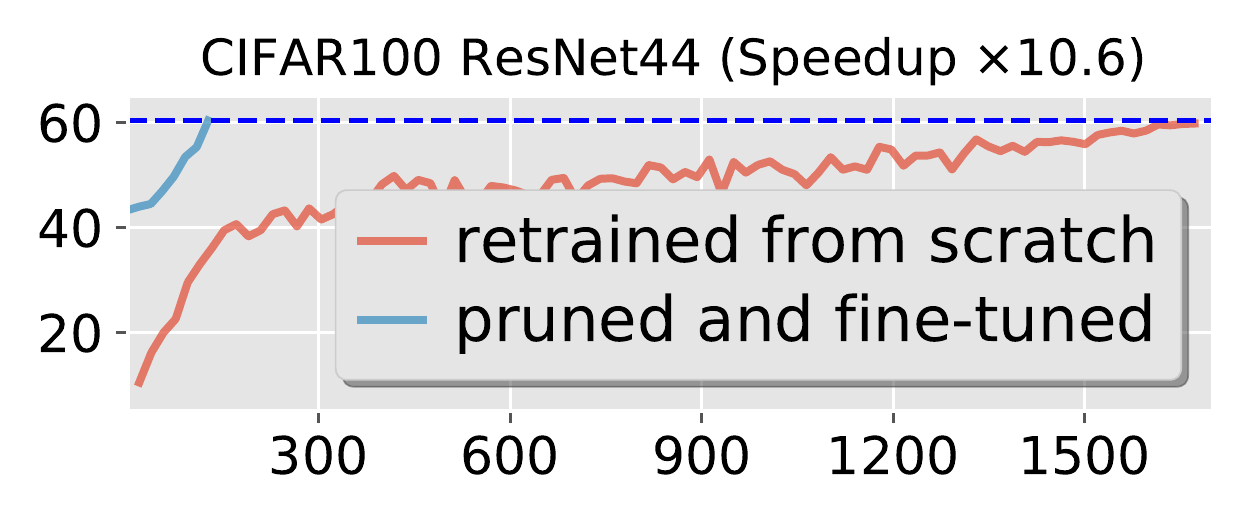}
    \includegraphics[width=0.23\columnwidth]{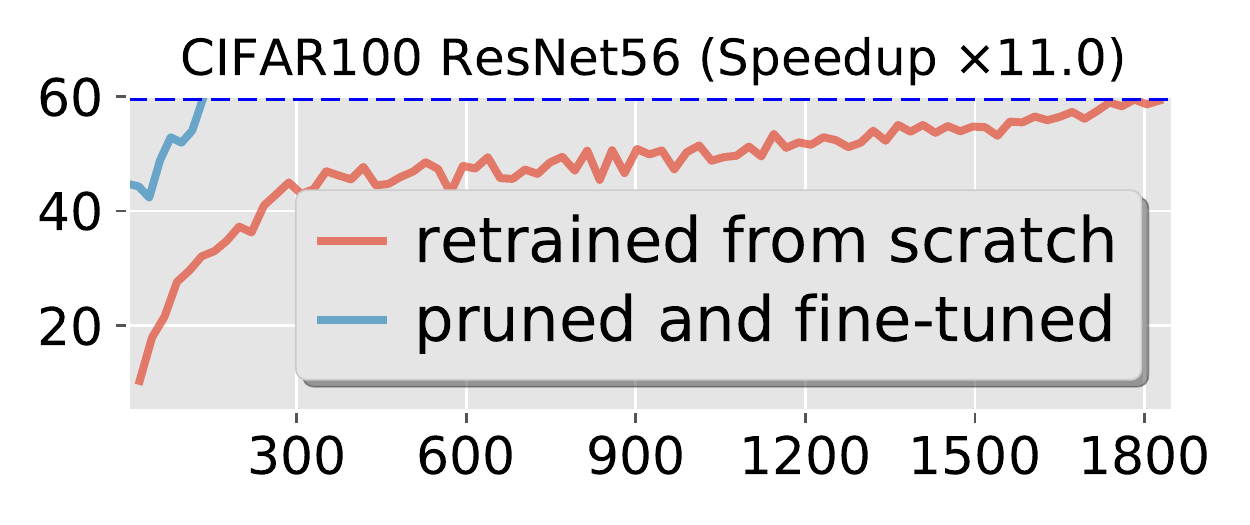}
  \end{minipage}
  \begin{minipage}[t]{1\linewidth}
    \centering
    \includegraphics[width=0.23\columnwidth]{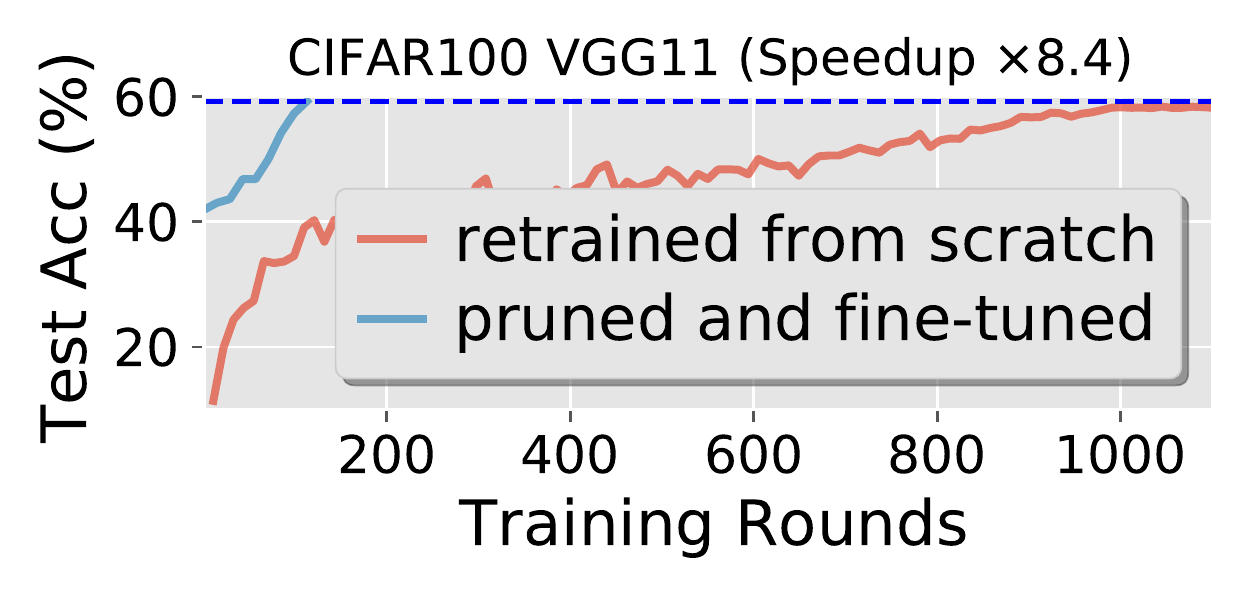}
    \includegraphics[width=0.23\columnwidth]{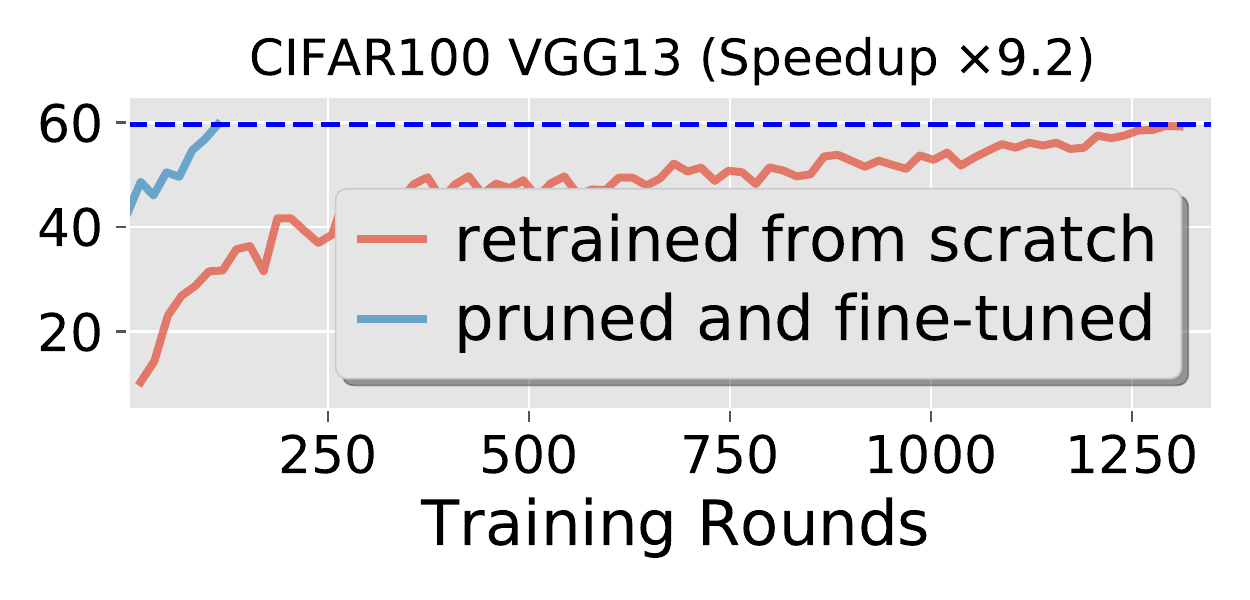}
    \includegraphics[width=0.23\columnwidth]{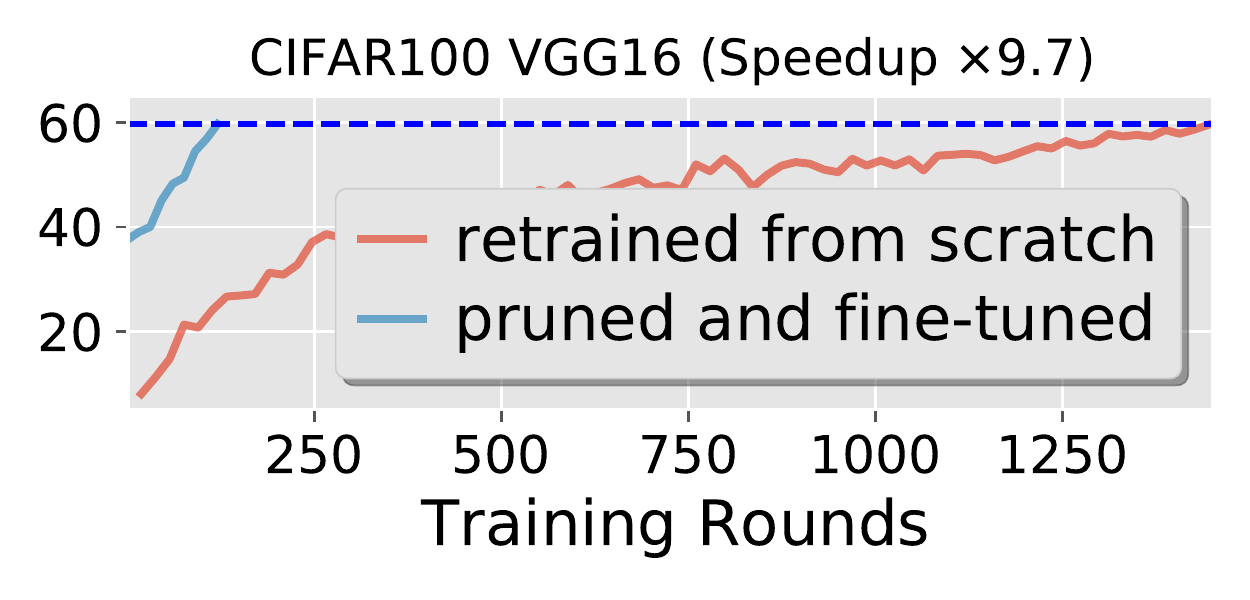}
    \includegraphics[width=0.23\columnwidth]{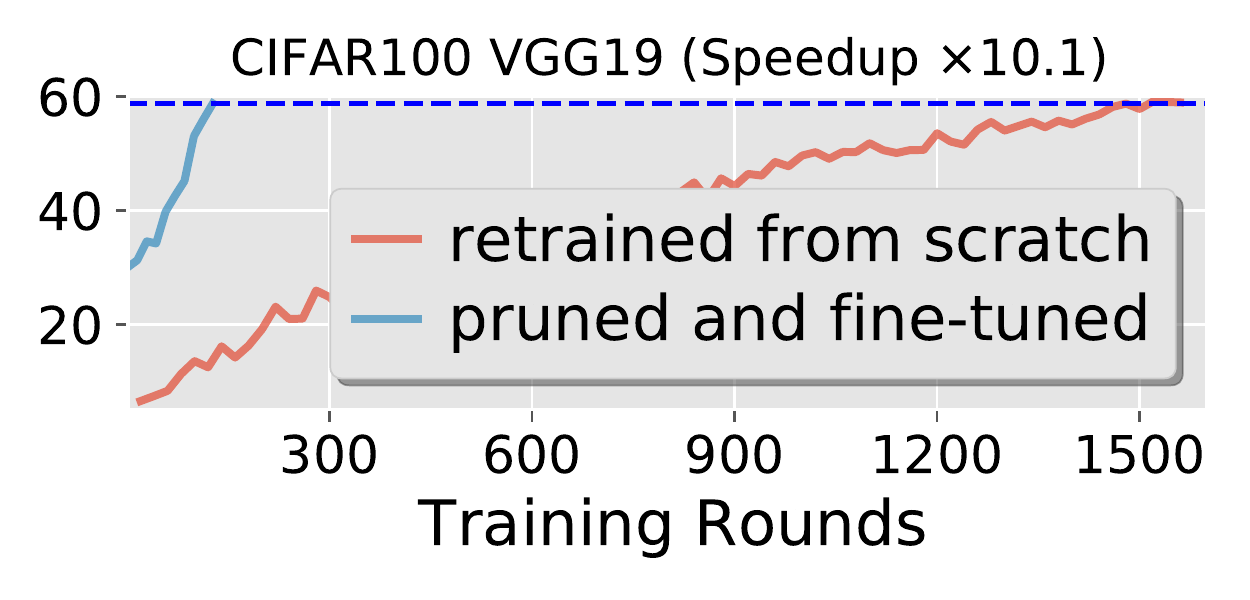}
  \end{minipage}
  \caption{Federated unlearning the last category of CIFAR10/CIFAR100 from pre-trained models.}
  \label{fig:retrain_time}
\end{figure*}

\begin{table*}[t]
\centering
\scriptsize 
\begin{tabular}{|c|c|c|c|c|c|c|c|c|c|c|c|c|}
\hline
&\multicolumn{6}{c|}{\textbf{CIFAR10}}&\multicolumn{6}{c|}{\textbf{CIFAR100}}\\
\cline{2-13}
&\multicolumn{3}{c|}{\textbf{Rounds of training}}&\multicolumn{3}{c|}{\textbf{Test accuracy on U/R-set}}&\multicolumn{3}{c|}{\textbf{Rounds of training}}&\multicolumn{3}{c|}{\textbf{Test accuracy on U/R-set}}\\
\hline
\textbf{Bias probability}
&\textbf{0.10}&\textbf{0.45}&\textbf{1.00}
&\textbf{0.10}&\textbf{0.45}&\textbf{1.00}
&\textbf{0.01}&\textbf{0.35}&\textbf{1.00}
&\textbf{0.01}&\textbf{0.35}&\textbf{1.00}\\
\hline
\textbf{Our method}&
113&135&181&
00.00/80.13\% & 00.00/74.45\% & 00.00/66.87\%&
110&163&235&
00.00/50.34\% & 00.00/46.99\% & 00.00/39.45\%
\\
\hline
\textbf{Fisher method}&
610&750&1110&
22.47/80.00\%&28.54/73.79\%&19.10/66.04\%&
700&820&1190&
15.33/49.86\%&14.71/45.30\%&17.09/38.32\%\\
\hline
\bottomrule
\end{tabular}
\caption{Impact of different bias probabilities (unlearning the last category from ResNet20 model).}
\label{tab:bias}
\vspace{-0.15in}
\end{table*}

We follow Fang et al. \cite{fang2020local} to distribute the training data points in a dataset among clients.
Assuming there are $M$ classes in a dataset.
We randomly split the clients into $M$ groups.
A training data point with label $m$ is assigned to group $m$ with probability $q$>0 and to any other group with probability (1-$q$)/($M$-1).
Within the same group, data are uniformly distributed to each client. 
$q$, as a bias probability, controls the distribution difference of the local training data.
If $q$=1/$M$, the local training data are independent and identically distributed (IID), otherwise the local training data are Non-IID.
A larger $q$ denotes a higher level of Non-IID among the local training data.
We will set $q$>1/$M$ to simulate the Non-IID settings.
Moreover, we assume the training data are distributed among 100 clients, and the number of clients participating in each round is 25. 
The bias probability $q$ is set from 0.1 to 1.0 for CIFAR10, and is set from 0.01 to 1.0 for CIFAR100.
On both CIFAR10 and CIFAR100, we compare our unlearning method against two baselines.

\textbf{Baselines}. (1) retraining the entire data (without the
points to be unlearned) from scratch; (2) forgetting the target category based on the Fisher unlearning method \cite{golatkar2020eternal,golatkar2020forgetting,golatkar2021mixed}. 
As introduced in Section~\ref{subsection:unlearning}, this method uses the Fisher information of the participant data and injects optimal noise in order to achieve unlearning.
\subsection{Experimental Results}
As shown in Figure~\ref{fig:retrain_time}, we unlearn the last category of CIFAR10 and CIFAR100 from different pre-trained models, and compare the speedup with fully retraining.
The learning method is \texttt{FedSGD}, learning rate is set 0.1, and bias probability is 0.1(0.01).
For full retrain, the x-axis represents the number of training rounds from scratch.
For our method, the x-axis represents the number of training rounds in fine-tuning process.
The y-axis represents the accuracy of the unlearned models on the testset (without the category to be forgotten). 
We show the impact of different bias probabilities in Table~\ref{tab:bias}. 
The learning method is \texttt{FedAVG}, local epoch is 5, learning rate is 0.1.
The model accuracy is calculated respectively on the testset only containing the target category (i.e., unlearned testset, or U-set) and the testset containing the rest categories (i.e., rest testset, or R-set). 
We set the hyper-parameter $R$=0.05 for CIFAR10 ResNet; 0.1 for CIFAR10 VGG; 0.1 for CIFAR100 ResNet; 0.15 for CIFAR100 VGG. 

\textbf{Unlearning speedup}. Obviously, the speedups produced by our method compared to fully retraining is very significant: on CIFAR10 dataset, we observe the speedup is 8.9$\sim$11.0× for the ResNet model, and 7.9$\sim$9.1× for the VGG model; on CIFAR100 dataset, the speedups are 9.9× and 8.4×, respectively.  
On the one hand, as the depth of the network increases, the effect of acceleration will get better.
On the other hand, as the total number of categories increases, the effect of unlearning acceleration will also become better. 
These results show that our method can be well adapted to real large-scale scenarios with a large number of categories and a deep network.

Compared to the baseline method using the Fisher information, our method performs significantly better because our method is insensitive to the distribution of training data, thereby can achieve a accurate unlearning even with severely biased participant data.
As another way to represent category information in the unlearning task, the Fisher method heavily depends on accurate and global access to all training data, and depicts a linear corrective Newton step to achieve unlearning by injecting optimal noise.
Once the training data cannot be globally accessed (i.e., with higher bias in our settings), the corrective step cannot be guided in precise. 
Therefore, a large value of noise will indiscriminately remove information from the model which in turn drastically reduces the effectiveness of further fine-tuned training.
Moreover, even the case of no bias, deep and nonlinear neural networks are great challenges for the linear corrective method.

\textbf{Information erasure}. As shown in Table~\ref{tab:erasure}, we compare the model accuracy on R/U-set respectively for our method and full retrain, which serves as a befitting manner to measure the quality of unlearning (i.e., assess how much information about the target category is still contained in the unlearned model).
It can be observed that, first of all, our method has no loss of precision on R-set.
On this basis, the accuracy of our method on U-set is the same as full retrain, and both are 0\%. 
These results prove that our method achieves resemble effect of information erasure on the pre-trained model as the full retrain does, without the expense of accuracy.

It is worth noting that, as the total number of categories increases, the accuracy of the after-pruned model on R-set is get lower (e.g., closing to 0\% for CIFAR100).
However, the speedup of unlearning is still kept as can be seen from Figure~\ref{fig:retrain_time}.
This proves from the side that our method can indeed cut the channels most relevant to the target class from the trained model.
In the fine-tuning process, less interference (on forward and backward propagation) comes from the channels related to the target category, thereby can effectively accelerate the recovery of accuracy on R-set.

\textbf{Multi-class removal}. As shown in Table~\ref{tab:multi}, the performance in the case of multi-class removal is still advanced (with the speedups comparable to the case of single-class removal).
After pruning for the second class, the accuracy of the after-pruned model on U-set gets lower (i.e., to 0\%), which shows an ideal information erasure.

\begin{table*}[t]
\centering
\tiny 
\begin{tabular}{|c|c|c|c|c|c|c|c|c|c|c|c|c|c|c|c|c|}
\hline
&\multicolumn{8}{c|}{\textbf{CIFAR10}}&\multicolumn{8}{c|}{\textbf{CIFAR100}}\\
\cline{2-17}
&\multicolumn{2}{c|}{\textbf{Raw model}}&\multicolumn{2}{c|}{\textbf{After-pruned}}&\multicolumn{2}{c|}{\textbf{Fine-tuned}}&\multicolumn{2}{c|}{\textbf{Re-trained}}&\multicolumn{2}{c|}{\textbf{Raw model}}&\multicolumn{2}{c|}{\textbf{After-pruned}}&\multicolumn{2}{c|}{\textbf{Fine-tuned}}&\multicolumn{2}{c|}{\textbf{Re-trained}}\\
\hline
\textbf{Accuracy}&\textbf{U-set}&\textbf{R-set}&\textbf{U-set}&\textbf{R-set}&\textbf{U-set}&\textbf{R-set}&\textbf{U-set}&\textbf{R-set}&\textbf{U-set}&\textbf{R-set}&\textbf{U-set}&\textbf{R-set}&\textbf{U-set}&\textbf{R-set}&\textbf{U-set}&\textbf{R-set}\\
\hline
\textbf{ResNet20}& 91.50\% & 83.33\% & 00.00\% & 20.79\% & 00.00\% & 86.40\% & 00.00\% & 86.33\% 
& 54.00\% & 50.01\% & 00.00\% & 05.38\% & 00.00\% & 57.17\% & 00.00\% & 57.11\% \\
\hline
\textbf{ResNet32}& 94.20\% & 83.71\% & 00.00\% & 11.58\% & 00.00\% & 86.40\% & 00.00\% & 86.14\% 
& 52.00\% & 51.67\% & 00.00\% & 01.06\% & 00.00\% & 59.62\% & 00.00\% & 59.42\% \\
\hline
\textbf{ResNet44}& 89.90\% & 83.94\% & 00.00\% & 22.19\% & 00.00\% & 86.48\% & 00.00\% & 86.34\%  
& 48.00\% & 53.25\% & 00.00\% & 01.22\% & 00.00\% & 60.41\% & 00.00\% & 59.85\% \\
\hline
\textbf{ResNet56}& 93.10\% & 84.02\% & 00.00\% & 11.11\% & 00.00\% & 86.42\% & 00.00\% & 86.38\% 
& 44.00\% & 52.91\% & 00.00\% & 01.32\% & 00.00\% & 59.60\% & 00.00\% & 59.28\% \\
\hline
\textbf{VGG11}& 88.20\% & 84.72\% & 00.00\% & 18.29\% & 00.00\% & 87.24\% & 00.00\% & 87.13\%
& 50.00\% & 53.55\% & 00.00\% & 01.28\% & 00.00\% & 59.25\% & 00.00\% & 58.20\% \\
\hline
\textbf{VGG13}& 91.50\% & 84.19\% & 00.00\% & 15.17\% & 00.00\% & 89.18\% & 00.00\% & 89.09\%
& 60.00\% & 51.88\% & 00.00\% & 03.82\% & 00.00\% & 59.65\% & 00.00\% & 59.27\% \\
\hline
\textbf{VGG16}& 91.60\% & 84.38\% & 00.00\% & 17.79\% & 00.00\% & 89.20\% & 00.00\% & 89.30\% 
& 44.00\% & 50.34\% & 00.00\% & 01.46\% & 00.00\% & 59.72\% & 00.00\% & 59.57\% \\
\hline
\textbf{VGG19}& 88.80\% & 83.53\% & 00.00\% & 11.11\% & 00.00\% & 89.72\% & 00.00\% & 89.62\%
& 52.00\% & 52.15\% & 00.00\% & 01.02\% & 00.00\% & 58.78\% & 00.00\% & 58.96\% \\
\hline
\bottomrule
\end{tabular}
\caption{Information forgotten effect compared to fully retraining (unlearning the last category).}
\label{tab:erasure}
\end{table*}

\begin{table*}[t]
\centering
\scriptsize 
\begin{tabular}{|c|c|c|c|c|c|c|c|c|c|c|c|}
\hline
&\multicolumn{2}{c|}{\textbf{Raw model}}&\multicolumn{2}{c|}{\textbf{First class pruned}}&\multicolumn{2}{c|}{\textbf{Second class pruned}}&\multicolumn{2}{c|}{\textbf{Fine-tuned}}&\multicolumn{3}{c|}{\textbf{Re-trained}}\\
\hline
\textbf{Model Accuracy}&\textbf{U-set}&\textbf{R-set}&\textbf{U-set}&\textbf{R-set}&\textbf{U-set}&\textbf{R-set}&\textbf{U-set}&\textbf{R-set}&\textbf{U-set}&\textbf{R-set}&\textbf{Speedup}
\\
\hline
\textbf{Delete class 0+9 from [0-9]}&
88.70\% & 83.01\% & 02.10\% & 24.57\% & 00.00\% & 32.41\% & 00.00\% & 87.12\% & 00.00\% & 87.29\% & ×8.71
\\
\hline
\textbf{Delete class 0+5 from [0-9]}&
81.90\% & 84.71\% & 00.25\% & 25.04\% & 00.00\% & 26.74\% & 00.00\% & 89.62\% & 00.00\% & 89.75\% &
×10.62\\
\hline
\textbf{Delete class 5+9 from [0-9]}&
84.70\% & 84.01\% & 02.10\% & 31.74\% & 00.00\% & 37.71\% & 00.00\% & 88.37\% & 00.00\% & 88.21\% & ×8.92
\\
\hline
\midrule
\end{tabular}
\caption{Delete two categories of CIFAR10 simultaneously from ResNet20 model.}
\label{tab:multi}
\end{table*}

\begin{table*}[t]
\scriptsize 
\centering
\begin{minipage}[t]{1.0\linewidth}
\centering
\begin{tabular}{|c|c|c|c|c|}
\hline
&\textbf{ResNet20}&\textbf{ResNet56}&\textbf{VGG11}&\textbf{VGG19}
\\
\hline
\textbf{Delete class 0 from [0-9]}&
00.35\% & 00.47\% & 00.39\% & 00.40\%
\\
\hline
\textbf{Delete class 9 from [0-9]}&
00.46\% & 00.50\% & 00.28\% & 00.37\%
\\
\hline
\textbf{Delete class 0+9 from [0-9]}&
00.51\% & 00.63\% & 00.49\% & 00.56\%
\\
\hline
\midrule
\end{tabular}
\caption{Difference of MIA success rate between fully retraining and our method under CIFAR10.}
\label{tab:mia}
\end{minipage}
\hspace{10pt}
\begin{minipage}[t]{1.0\linewidth}
\centering
\begin{tabular}{|c|c|c|c|}
\hline
&\textbf{ResNet20}&\textbf{ResNet56}&\textbf{VGG19}
\\
\hline
 \textbf{Delete class 0 from [0-9]} & 00.33\%/3.01e-6 & 00.46\%/4.50e-6 & 00.35\%/3.65e-6 
\\
\hline
 \textbf{Delete class 9 from [0-9]} & 00.27\%/2.58e-6 & 00.43\%/4.84e-6 & 00.32\%/3.87e-6
\\
\hline
 \textbf{Delete class 0+9 from [0-9]} & 00.36\%/2.63e-6 & 00.59\%/5.27e-6 & 00.41\%/3.55e-6
\\
\hline
\midrule
\end{tabular}
\caption{Distance of classification accuracy distribution between fully retraining and our method under CIFAR10.}
\label{tab:dist}
\end{minipage}
\end{table*}

\textbf{Membership inference attack}. 
We launch a membership inference attack (MIA) introduced in literature \cite{shokri2017membership}.
Given a sample of target category and the model after unlearning, the attack aims to infer whether the sample was used to train the original model.
In other words, the attack aims to know that the target sample is in the training dataset of the original model.
While the goal of unlearning is to protect privacy information about the target sample, a successful attack considered here can show unlearning instead leaks the target sample's privacy.
We measure the success rate of attacks to evaluate how much information about the target category is still contained in the model after unlearning.
Seen from results under CIFAR10 (as referred to Table~\ref{tab:mia}), no significant difference is found in the attack success rate between fully retraining and our method (with average deviation less than 0.63\%).

\textbf{Classification accuracy distribution}.
We investigate how are the classification accuracies distributed among all classes after unlearning and whether it is an ideal distribution approximate that of retraining from scratch.
As shown in Table~\ref{tab:dist}, the class-wise accuracy difference is only 0.59\% at most.
Moreover, we use the KL divergence to evaluate the distance between our distribution and the ideal distribution.
Seen from results under CIFAR10, the KL divergence between fully retraining and our method is very close to 0 (less than 5.27e-6).

\section{Related Work}
Machine unlearning as the field most related to this paper, its notion was first proposed by Cao et al. \cite{cao2015towards} in the context of statistical query learning.
Different from the class-level unlearning we defined, the goal of machine unlearning is more strict as it forgets a particular subset of samples within a class. 
To unlearn a data sample, Cao et al. \cite{cao2015towards} transforms the learning algorithms into a summation form that follows statistical query learning, breaking down the dependencies of training data.
However, the method in \cite{cao2015towards} is not applicable to learning algorithms that cannot be
transformed into summation form, such as neural networks.

Thereafter, machine unlearning for different ML models have been explored.
Ginart et al. \cite{ginart2019making} studied the problem of data removal for variants of k-means clustering, but cannot be applied to supervised learning.
The works \cite{golatkar2020eternal,golatkar2020forgetting,golatkar2021mixed,guo2020certified,izzo2021approximate,wu2020deltagrad,wu2020priu,graves2021amnesiac,neel2021descent} introduced in Section~\ref{subsection:unlearning} focus on supervised linear classification models, such as logistic regression, but not specializing for more complex models, such as neural networks.
There are other approximate unlearning methods that fit for specific ML models, such as Brophy et al. \cite{brophy2021machine} for random forest models, or Nguyen et al. \cite{nguyen2020variational} for bayesian learning.
Bourtoule et al. \cite{bourtoule2021machine} proposed a more general algorithm named \texttt{SISA}.
The main idea of \texttt{SISA} is to split the training data into several disjoint shards, with each shard training one sub-model. 
To remove a specific sample, the algorithm only needs to retrain the sub-model that contains this sample. 
Nevertheless, existing machine unlearning works focus on ML models in traditional centralized settings, i.e., the training data assumed to be globally accessed, which is ineligible for unlearning in the FL settings. %
Liu et al. \cite{liu2021federaser} studied client-level unlearning in FL, i.e., to eliminate the influence of data of a FL device on the global model, and proposed an algorithm named \texttt{FedEraser}.
The main idea of \texttt{FedEraser} is to unlearn model by leveraging the historical parameter updates of clients that have been retained at the federated server during the training process.
However, for convolutional architectures, \cite{liu2021federaser} show that unlearning is possible for only shallow networks, such as for a 2-layer CNN followed by 2 fully-connected (FC) layers.
How much the amount of stored information in the federated server also leads to a trade-off between the accuracy and scalability.

\section{Conclusion}
We propose a definition of selectively forgetting categories for the trained CNNs classification model in the federated learning (FL) settings.
Based on this definition, we propose a scrubbing procedure that cleans information about particular categories from the trained model, without the need to globally access the training data, nor to retrain from scratch.
Compared to the gold standard model trained from scratch without any knowledge of the target category data, our method demonstrates its superiority in terms of unlearning speed under no degradation in accuracy. 
We envision this work as an early step towards federated unlearning, which contributes as a complementary block for compliance with legal and ethical criteria.
In future work, we will investigate a more challenging fine-grained unlearning task in FL settings, where not only does an entire class need to be forgotten, but rather a particular subset of samples within a class needs to be removed, while still maintaining output knowledge of that class.

\bibliographystyle{unsrt}  
\bibliography{main}  

\begin{thebibliography}{10}

\bibitem{zhao2018federated}
Yue Zhao, Meng Li, Liangzhen Lai, Naveen Suda, Damon Civin, and Vikas Chandra.
\newblock Federated learning with non-iid data.
\newblock {\em arXiv preprint arXiv:1806.00582}, 2018.

\bibitem{kairouz2019advances}
Peter Kairouz, H~Brendan McMahan, Brendan Avent, Aur{\'e}lien Bellet, Mehdi
  Bennis, Arjun~Nitin Bhagoji, Kallista Bonawitz, Zachary Charles, Graham
  Cormode, Rachel Cummings, et~al.
\newblock Advances and open problems in federated learning.
\newblock {\em arXiv preprint arXiv:1912.04977}, 2019.

\bibitem{mcmahan2017communication}
Brendan McMahan, Eider Moore, Daniel Ramage, Seth Hampson, and Blaise~Aguera
  y~Arcas.
\newblock Communication-efficient learning of deep networks from decentralized
  data.
\newblock In {\em Proceedings of International Conference on Artificial
  Intelligence and Statistics (AISTATS)}, 2017.

\bibitem{yang2019federated}
Qiang Yang, Yang Liu, Tianjian Chen, and Yongxin Tong.
\newblock Federated machine learning: Concept and applications.
\newblock {\em ACM Transactions on Intelligent Systems and Technology (TIST)},
  10(2):1--19, 2019.

\bibitem{ginart2019making}
Antonio Ginart, Melody~Y Guan, Gregory Valiant, and James Zou.
\newblock Making ai forget you: Data deletion in machine learning.
\newblock In {\em Proceedings of Conference on Neural Information Processing
  Systems (NeurIPS)}, 2019.

\bibitem{bourtoule2021machine}
Lucas Bourtoule, Varun Chandrasekaran, Christopher~A Choquette-Choo, Hengrui
  Jia, Adelin Travers, Baiwu Zhang, David Lie, and Nicolas Papernot.
\newblock Machine unlearning.
\newblock In {\em Proceedings of IEEE Symposium on Security and Privacy (SP)},
  2021.

\bibitem{guo2020certified}
Chuan Guo, Tom Goldstein, Awni Hannun, and Laurens Van Der~Maaten.
\newblock Certified data removal from machine learning models.
\newblock In {\em Proceedings of International Conference on Machine Learning
  (ICML)}, 2020.

\bibitem{golatkar2020eternal}
Aditya Golatkar, Alessandro Achille, and Stefano Soatto.
\newblock Eternal sunshine of the spotless net: Selective forgetting in deep
  networks.
\newblock In {\em Proceedings of IEEE/CVF Conference on Computer Vision and
  Pattern Recognition (CVPR)}, 2020.

\bibitem{golatkar2020forgetting}
Aditya Golatkar, Alessandro Achille, and Stefano Soatto.
\newblock Forgetting outside the box: Scrubbing deep networks of information
  accessible from input-output observations.
\newblock In {\em Proceedings of European Conference on Computer Vision
  (ECCV)}, 2020.

\bibitem{golatkar2021mixed}
Aditya Golatkar, Alessandro Achille, Avinash Ravichandran, Marzia Polito, and
  Stefano Soatto.
\newblock Mixed-privacy forgetting in deep networks.
\newblock In {\em Proceedings of IEEE/CVF Conference on Computer Vision and
  Pattern Recognition (CVPR)}, 2021.

\bibitem{wu2020deltagrad}
Yinjun Wu, Edgar Dobriban, and Susan Davidson.
\newblock Deltagrad: Rapid retraining of machine learning models.
\newblock In {\em Proceedings of International Conference on Machine Learning
  (ICML)}, 2020.

\bibitem{wu2020priu}
Yinjun Wu, Val Tannen, and Susan~B Davidson.
\newblock Priu: A provenance-based approach for incrementally updating
  regression models.
\newblock In {\em Proceedings of ACM SIGMOD International Conference on
  Management of Data}, 2020.

\bibitem{mantelero2013eu}
Alessandro Mantelero.
\newblock The eu proposal for a general data protection regulation and the
  roots of the ‘right to be forgotten’.
\newblock {\em Computer Law \& Security Review}, 29(3):229--235, 2013.

\bibitem{shastri2019seven}
Supreeth Shastri, Melissa Wasserman, and Vijay Chidambaram.
\newblock The seven sins of personal-data processing systems under
  $\{$GDPR$\}$.
\newblock In {\em Proceedings of USENIX Workshop on Hot Topics in Cloud
  Computing (HotCloud)}, 2019.

\bibitem{martens2020new}
James Martens.
\newblock New insights and perspectives on the natural gradient method.
\newblock {\em Journal of Machine Learning Research}, 21(146):1--76, 2020.

\bibitem{izzo2021approximate}
Zachary Izzo, Mary~Anne Smart, Kamalika Chaudhuri, and James Zou.
\newblock Approximate data deletion from machine learning models.
\newblock In {\em Proceedings of International Conference on Artificial
  Intelligence and Statistics (AISTATS)}, 2021.

\bibitem{koh2017understanding}
Pang~Wei Koh and Percy Liang.
\newblock Understanding black-box predictions via influence functions.
\newblock In {\em Proceedings of International Conference on Machine Learning
  (ICML)}, 2017.

\bibitem{graves2021amnesiac}
Laura Graves, Vineel Nagisetty, and Vijay Ganesh.
\newblock Amnesiac machine learning.
\newblock In {\em Proceedings of the AAAI Conference on Artificial
  Intelligence}, 2021.

\bibitem{neel2021descent}
Seth Neel, Aaron Roth, and Saeed Sharifi-Malvajerdi.
\newblock Descent-to-delete: Gradient-based methods for machine unlearning.
\newblock In {\em Proceedings of International Conference on Algorithmic
  Learning Theory (ALT)}, 2021.

\bibitem{yosinski2015understanding}
Jason Yosinski, Jeff Clune, Anh Nguyen, Thomas Fuchs, and Hod Lipson.
\newblock Understanding neural networks through deep visualization.
\newblock {\em arXiv preprint arXiv:1506.06579}, 2015.

\bibitem{paik2013novel}
Jiaul~H Paik.
\newblock A novel tf-idf weighting scheme for effective ranking.
\newblock In {\em Proceedings of ACM SIGIR Conference on Research and
  Development in Information Retrieval}, 2013.

\bibitem{yahav2018comments}
Inbal Yahav, Onn Shehory, and David Schwartz.
\newblock Comments mining with tf-idf: the inherent bias and its removal.
\newblock {\em IEEE Transactions on Knowledge and Data Engineering (TKDE)},
  31(3):437--450, 2018.

\bibitem{he2017channel}
Yihui He, Xiangyu Zhang, and Jian Sun.
\newblock Channel pruning for accelerating very deep neural networks.
\newblock In {\em Proceedings of IEEE International Conference on Computer
  Vision (ICCV)}, 2017.

\bibitem{lin2020channel}
Mingbao Lin, Rongrong Ji, Yuxin Zhang, Baochang Zhang, Yongjian Wu, and
  Yonghong Tian.
\newblock Channel pruning via automatic structure search.
\newblock In {\em Proceedings of International Joint Conference on Artificial
  Intelligence (IJCAI)}, 2020.

\bibitem{kung2019methodical}
Sun-Yuan Kung, Zejiang Hou, and Yuchen Liu.
\newblock Methodical design and trimming of deep learning networks: Enhancing
  external bp learning with internal omnipresent-supervision training paradigm.
\newblock In {\em Proceedings of IEEE International Conference on Acoustics,
  Speech and Signal Processing (ICASSP)}, 2019.

\bibitem{zhuang2018discrimination}
Zhuangwei Zhuang, Mingkui Tan, Bohan Zhuang, Jing Liu, Yong Guo, Qingyao Wu,
  Junzhou Huang, and Jinhui Zhu.
\newblock Discrimination-aware channel pruning for deep neural networks.
\newblock In {\em Proceedings of Conference on Neural Information Processing
  Systems (NeurIPS)}, 2018.

\bibitem{nair2010rectified}
Vinod Nair and Geoffrey~E Hinton.
\newblock Rectified linear units improve restricted boltzmann machines.
\newblock In {\em Proceedings of International Conference on Machine Learning
  (ICML)}, 2010.

\bibitem{krizhevsky2009learning}
Alex Krizhevsky, Geoffrey Hinton, et~al.
\newblock Learning multiple layers of features from tiny images.
\newblock 2009.

\bibitem{he2016deep}
Kaiming He, Xiangyu Zhang, Shaoqing Ren, and Jian Sun.
\newblock Deep residual learning for image recognition.
\newblock In {\em Proceedings of IEEE/CVF conference on Computer Vision and
  Pattern Recognition}, 2016.

\bibitem{simonyan2014very}
Karen Simonyan and Andrew Zisserman.
\newblock Very deep convolutional networks for large-scale image recognition.
\newblock {\em arXiv preprint arXiv:1409.1556}, 2014.

\bibitem{konevcny2016federated}
Jakub Kone{\v{c}}n{\`y}, H~Brendan McMahan, Felix~X Yu, Peter Richt{\'a}rik,
  Ananda~Theertha Suresh, and Dave Bacon.
\newblock Federated learning: Strategies for improving communication
  efficiency.
\newblock {\em arXiv preprint arXiv:1610.05492}, 2016.

\bibitem{fang2020local}
Minghong Fang, Xiaoyu Cao, Jinyuan Jia, and Neil Gong.
\newblock Local model poisoning attacks to byzantine-robust federated learning.
\newblock In {\em Proceedings of USENIX Security Symposium (Security)}, 2020.

\bibitem{shokri2017membership}
Reza Shokri, Marco Stronati, Congzheng Song, and Vitaly Shmatikov.
\newblock Membership inference attacks against machine learning models.
\newblock In {\em Proceedings of IEEE Symposium on Security and Privacy (SP)},
  pages 3--18, 2017.

\bibitem{cao2015towards}
Yinzhi Cao and Junfeng Yang.
\newblock Towards making systems forget with machine unlearning.
\newblock In {\em Proceedings of IEEE Symposium on Security and Privacy (SP)},
  2015.

\bibitem{brophy2021machine}
Jonathan Brophy and Daniel Lowd.
\newblock Machine unlearning for random forests.
\newblock In {\em Proceedings of International Conference on Machine Learning
  (ICML)}, 2021.

\bibitem{nguyen2020variational}
Quoc~Phong Nguyen, Bryan Kian~Hsiang Low, and Patrick Jaillet.
\newblock Variational bayesian unlearning.
\newblock In {\em Proceedings of Conference on Neural Information Processing
  Systems (NeurIPS)}, 2020.

\bibitem{liu2021federaser}
Gaoyang Liu, Xiaoqiang Ma, Yang Yang, Chen Wang, and Jiangchuan Liu.
\newblock Federaser: Enabling efficient client-level data removal from
  federated learning models.
\newblock In {\em Proceedings of IEEE/ACM International Symposium on Quality of
  Service (IWQoS)}, 2021.

\end{thebibliography}

\clearpage

\appendix
\section{Appendices}
\subsection{Multi-class removal: extensive cases}
\begin{figure*}[t]
  \begin{minipage}[t]{1\linewidth}
    \centering
    \includegraphics[width=0.23\columnwidth]{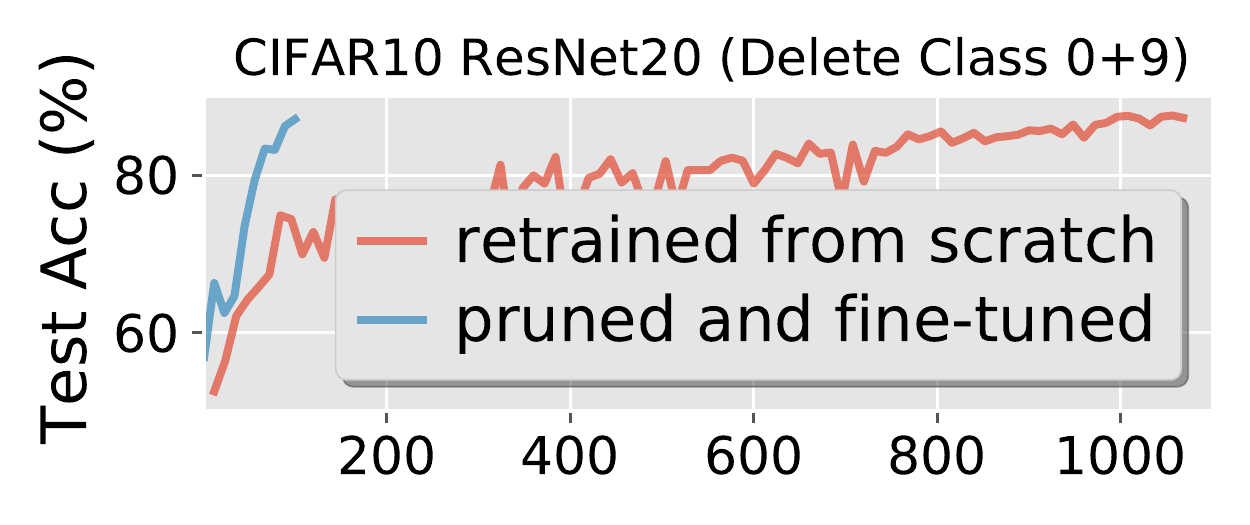}
    \includegraphics[width=0.23\columnwidth]{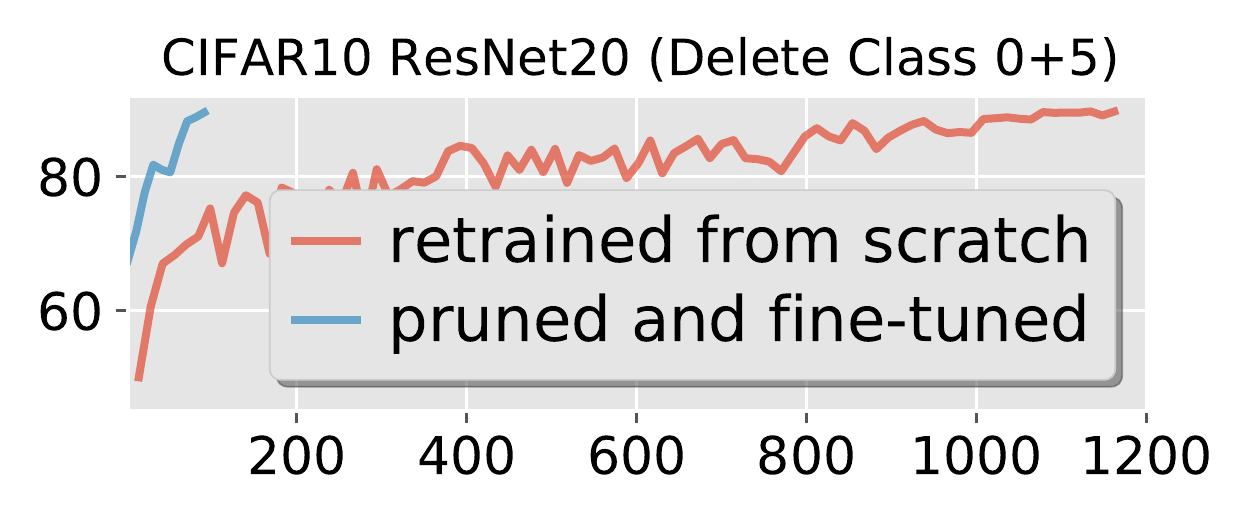}
    \includegraphics[width=0.23\columnwidth]{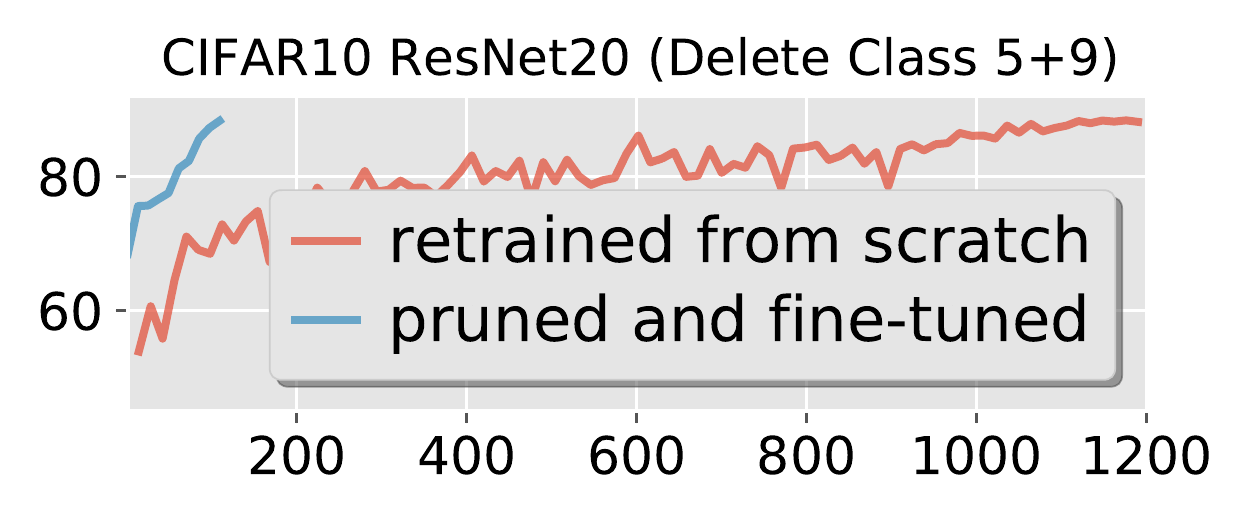}
    \includegraphics[width=0.23\columnwidth]{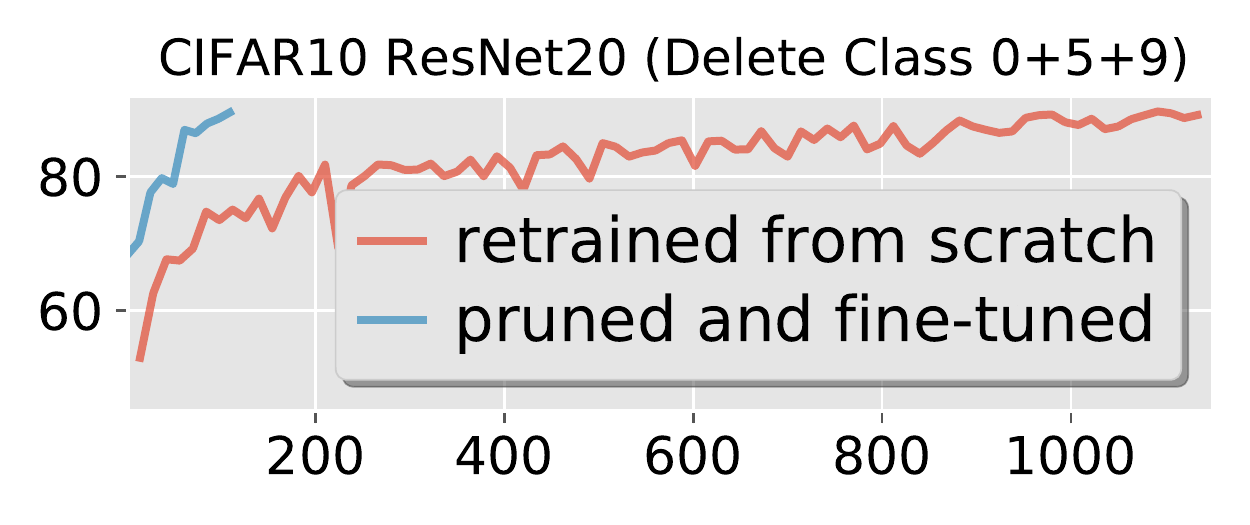}
  \end{minipage}
  \begin{minipage}[t]{1\linewidth}
    \centering
    \includegraphics[width=0.23\columnwidth]{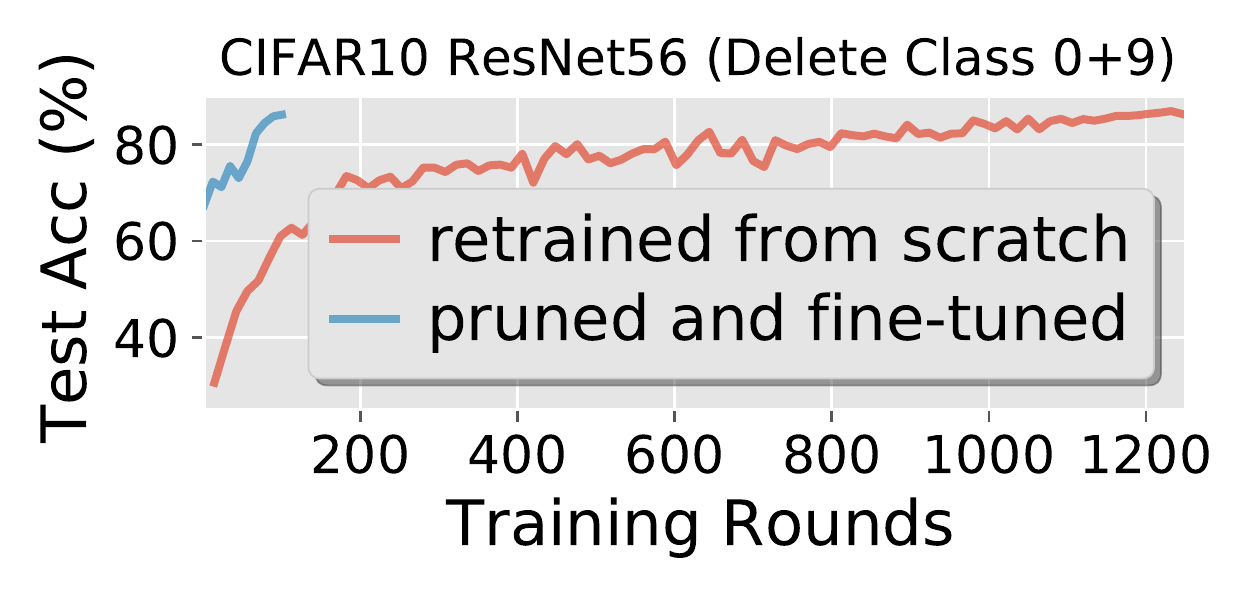}
    \includegraphics[width=0.23\columnwidth]{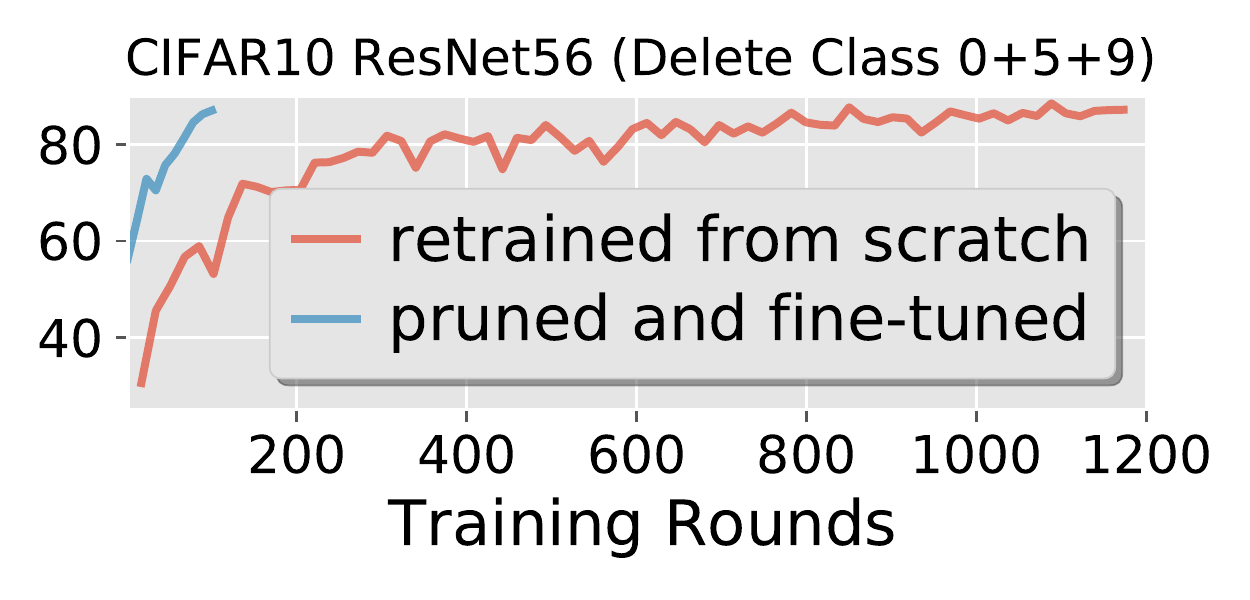}
    \includegraphics[width=0.23\columnwidth]{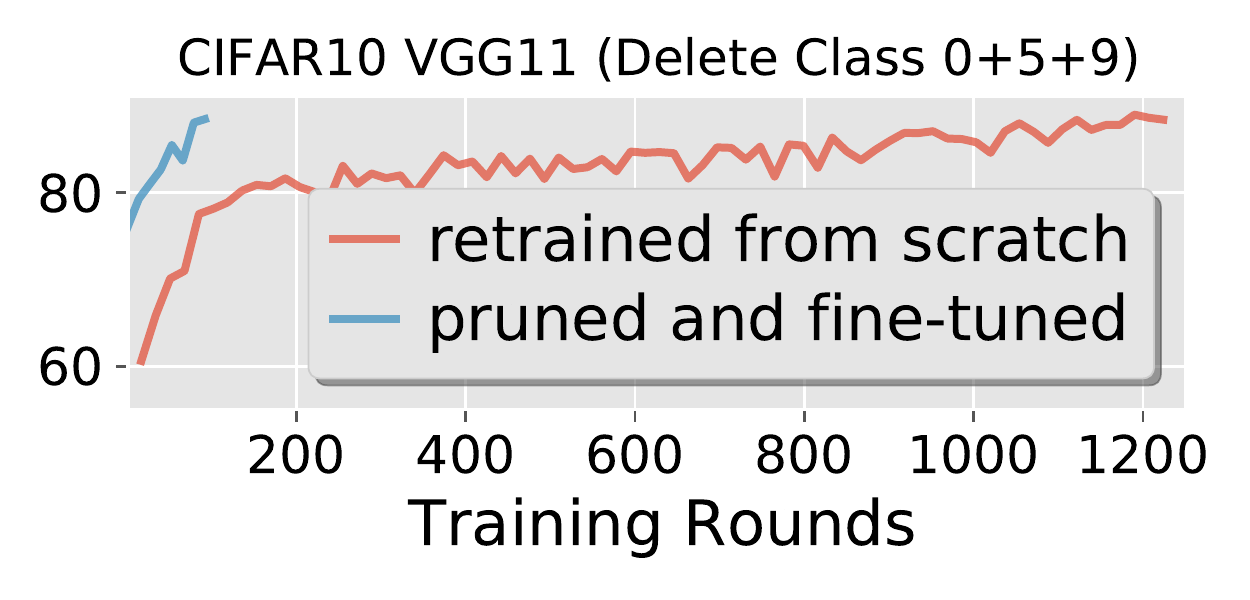}
    \includegraphics[width=0.23\columnwidth]{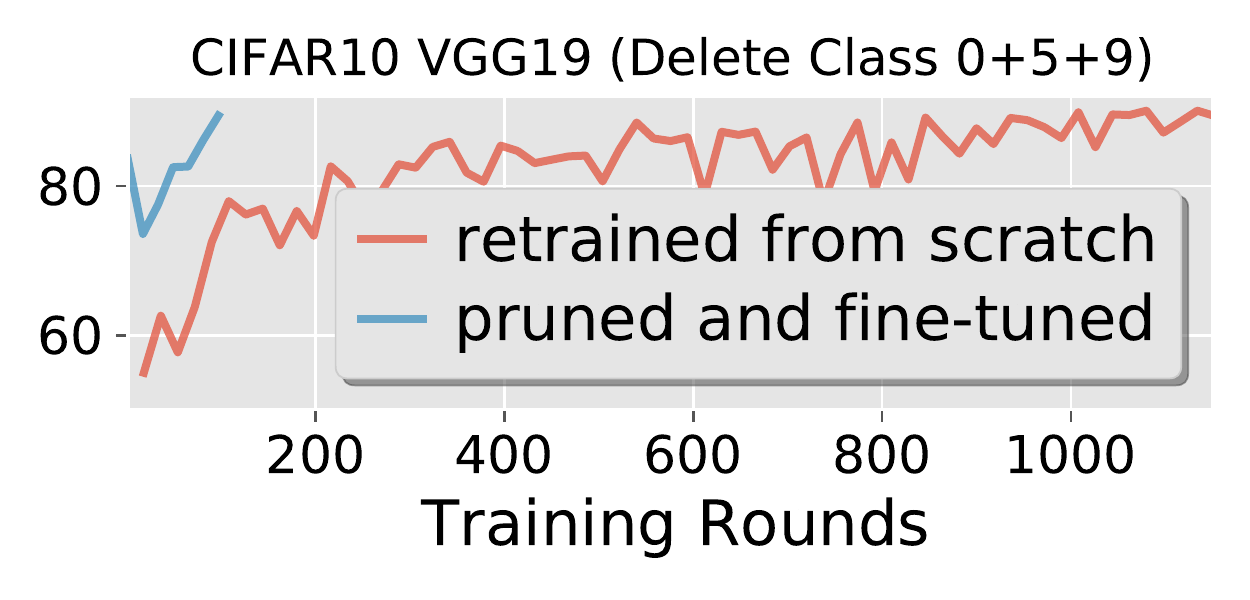}
  \end{minipage}
  \caption{Extensive experiments: removing multiple categories of CIFAR10 from pre-trained models.}
  \label{fig:multi}
\end{figure*}

In the case of multi-class removal, Table~\ref{tab:multi} has shown the effect of deleting two categories of CIFAR10 simultaneously from ResNet20.
For a more comprehensive investigation, we conduct similar experiments on CIFAR10 dataset with more pre-trains models (ResNet20/56, VGG11/19), as shown in Figure~\ref{fig:multi} and Table~\ref{tab:multi_full}. 
The learning method is \texttt{FedSGD}, learning rate is set 0.1, and bias probability is 0.1(0.01).
As seen, the performance of our method in unlearning acceleration and information erasure is satisfactory. 

\subsection{Term frequency inverse document frequency (TF-IDF): formal definition}
As introduced in Section~\ref{subsec:tf-idf}, TF-IDF for a word in a document is quantized by multiplying TF and IDF.
TF-IDF multiplies the frequency of term (TF) appears in a document and the inverse document frequency (IDF) of the term across a collection of documents.
In particular, TF-IDF score for a term $\bm{t}$ in document $\bm{e}$ from the document set $\bm{E}$=\{$\bm{e}_1$,$\bm{e}_2$,$\dots$,$\bm{e}_n$\} is calculated:
\begin{align}\label{equ:app1}
    \text{TF-IDF}(\bm{t},\bm{e},\bm{E})=\text{TF}(\bm{t},\bm{e})*\text{IDF}(\bm{t},\bm{E})
\end{align}
where TF(·) and IDF(·) are defined to calculate as:
\begin{align}\label{equ:app2}
\text{TF}(\bm{t},\bm{e})&=\texttt{freq}(\bm{t},\bm{e})/|\bm{e}|\\
\text{IDF}(\bm{t},\bm{E})=\log[(|\bm{E}|+1)&/(\texttt{count}(\bm{e},\bm{e}\in\bm{E}\cap\bm{t}\in\bm{e})+1)]
\end{align}

TF equals that the number of occurrences $\texttt{freq}(\bm{t},\bm{e})$ in document $\bm{e}$ divided by the length $|\bm{e}|$ of this document, which eliminates the impact of the length of the document on the term frequency.
IDF equals the length of the set $\bm{E}$ of all documents, divided by the length $\texttt{count}(\bm{e},\bm{e}\in\bm{E}\cap\bm{t}\in\bm{e})$ of the subset of documents term $\bm{t}$ appears in each of these documents.

\subsection{\texttt{ReLU} activation and \texttt{avg\_pool2d} operation: more details in Section~\ref{subsec:local_proc}}
At each client, \texttt{local\_proc} in the unlearning program is conducted with local private images, in order to generate a local representation between channels and categories, and facilitate measuring the class-wise discrimination of different channels.

As a first step, the client will record the feature map generated by local model in each layer.
Then, for $l$-th layer, we reduce the dimension of feature map tensor $\bm{O}_l^i$ of $i$-th category image by a ReLU layer and an average pool layer as:
\begin{align}\label{equ:app3}
    {\bm{A}'}_l^i=\left[\sum_1^{H_{l+1}}\sum_1^{W_{l+1}}\max(0,\bm{O}_l^i)\right]/(H_{l+1}*W_{l+1})
\end{align} 
where $\bm{O}_l^i\in\mathcal{R}^{C_{in}^{l+1}\times H_{l+1} \times W_{l+1}}$, and $C_{in}^{l+1}$=$C_{out}^l$.
We refer ${\bm{A}'}_l^i\in\mathcal{R}^{C_{out}^{l}\times 1}$ as the category
activation value of channels, and we denote the local representation between classes and channels as  
\begin{align}\label{equ:app4}
\bm{A}'_l=\left\{{\bm{A}'}_l^1, {\bm{A}'}_l^2, \dots, {\bm{A}'}_l^{|\bm{U}|}\right\}\in\mathcal{R}^{|\bm{U}|\times C_{out}^{l}}
\end{align} 
The global representation ${\bm{A}^*}_l^i$ of $i$-th category is calculated by averaging local representation $\bm{A}'_l$ for all $i$-th class images. 
we denote the global representation between classes and channels as

\begin{align}\label{equ:app5}
\bm{A}^*_l=\left\{{\bm{A}^*}_l^1, {\bm{A}^*}_l^2, \dots, {\bm{A}^*}_l^{|\bm{U}|}\right\}\in\mathcal{R}^{|\bm{U}|\times C_{out}^{l}}
\end{align} 

\subsection{Algorithm of the framework}
The procedure steps in the framework (as referred to Section~\ref{subsec:framework}) can be found from Algorithm~\ref{alg:framework}.

\subsection{Algorithm of \texttt{local\_proc}}
The procedure steps in \texttt{local\_proc} (as referred to Section~\ref{subsec:local_proc}) can be found from Algorithm~\ref{alg:local-proc}.
\begin{algorithm}[t] 
\caption{Procedure steps in \texttt{local\_proc}.} 
\label{alg:local-proc} 
\begin{algorithmic}[1]
\REQUIRE
the collection of local private images, $\bm{X}_1$;\\
the local FL model that has been pre-trained, $\bm{w}$;
\ENSURE
local representation between classes and channels, $\bm{A}'$;
\STATE the client conducts the unlearning program in its local;\\
\FOR{each model layer $l$ in $L$}
\STATE record output feature map in $l$-th layer $\bm{O}_l$=$\bm{X}_l\circledast \bm{w}_l$;\\
\STATE $\bm{O}'_l\leftarrow$ apply \texttt{ReLU} activation to output feature map $\bm{O}_l$;\\
\STATE $\bm{A}_l\leftarrow$ apply \texttt{avg\_pool2d} to after-activated $\bm{O}'_l$;\\
\STATE $\bm{A}'_l\leftarrow$ impose $\bm{A}_l$ to be averaged by class-wise;
\ENDFOR
\RETURN $\bm{A}'$=$\{\bm{A}'_1, \bm{A}'_2, \dots, \bm{A}'_L\}$;
\end{algorithmic}
\end{algorithm}

\subsection{Algorithm of \texttt{server\_proc}}
The procedure steps in \texttt{server\_proc} (as referred to Section~\ref{subsec:server_proc}) can be found from Algorithm~\ref{alg:server-proc}.
\begin{algorithm}[t] 
\caption{Procedure steps in \texttt{server\_proc}.} 
\label{alg:server-proc} 
\begin{algorithmic}[1]
\REQUIRE
the class to be forgotten, $u'$;\\
the set of total available classes, $\bm{U}$;\\
the collection of participant FL clients, $\bm{n}$;\\
the global FL model that has been pre-trained, $\bm{w}$;\\
local representations between classes and channels, $\bm{A}'$;\\
hyper-parameter to determine threshold of pruning $R$;
\ENSURE
FL model with discriminative channels pruned, $\hat{\bm{w}}$;
\STATE $\bm{A}^*\leftarrow$ aggregate $\bm{A}'_i$ in average for each $i\in\bm{n}$;\\ 
\FOR{each model layer $l$ in $L$}
\STATE calculate TF$_l$($\bm{A}^*$, $u'$), IDF$_l$($\bm{A}^*$, $\bm{U}$), TF-IDF$_l$(TF$_l$, IDF$_l$)\\
/* according to Equation~(\ref{equ:tf}), (\ref{equ:idf}) and (\ref{equ:tf-idf}) */\\
\STATE TF-IDF $\leftarrow$\texttt{concatnate}(TF-IDF$_l$);
\ENDFOR
\STATE $\hat{\bm{w}}\leftarrow$\texttt{channel\_pruner}($\bm{w}$, TF-IDF, $R$);
\RETURN $\hat{\bm{w}}$;
\end{algorithmic}
\end{algorithm}

\begin{table*}[t]
\centering
\scriptsize
\begin{tabular}{|c|c|c|c|c|c|c|c|c|c|c|c|}
\hline
ResNet20 CIFAR10 &\multicolumn{2}{c|}{\textbf{Raw model}}&\multicolumn{2}{c|}{\textbf{First class pruned}}&\multicolumn{2}{c|}{\textbf{Last class pruned}}&\multicolumn{2}{c|}{\textbf{Fine-tuned}}&\multicolumn{3}{c|}{\textbf{Re-trained}}\\
\hline
\textbf{Model Accuracy}&\textbf{U-set}&\textbf{R-set}&\textbf{U-set}&\textbf{R-set}&\textbf{U-set}&\textbf{R-set}&\textbf{U-set}&\textbf{R-set}&\textbf{U-set}&\textbf{R-set}&\textbf{Speedup}
\\
\hline
\textbf{Delete class 0+9 from [0-9]}&
88.70\% & 83.01\% & 02.10\% & 24.57\% & 00.00\% & 32.41\% & 00.00\% & 87.12\% & 00.00\% & 87.29\% & ×8.71
\\
\hline
\textbf{Delete class 0+5 from [0-9]}&
81.90\% & 84.71\% & 00.25\% & 25.04\% & 00.00\% & 26.74\% & 00.00\% & 89.62\% & 00.00\% & 89.75\% &
×10.62\\
\hline
\textbf{Delete class 5+9 from [0-9]}&
84.70\% & 84.01\% & 02.10\% & 31.74\% & 00.00\% & 37.71\% & 00.00\% & 88.37\% & 00.00\% & 88.21\% & ×8.92
\\
\hline
\textbf{Delete class 0+5+9 from [0-9]}&
85.10\% & 83.74\% & 01.57\% & 28.01\% & 00.00\% & 30.00\% & 00.00\% & 89.62\% & 00.00\% & 89.23\% & ×8.45
\\
\hline
\bottomrule
\end{tabular}
\begin{tabular}{|c|c|c|c|c|c|c|c|c|c|c|c|}
\hline
ResNet56 CIFAR10 &\multicolumn{2}{c|}{\textbf{Raw model}}&\multicolumn{2}{c|}{\textbf{First class pruned}}&\multicolumn{2}{c|}{\textbf{Last class pruned}}&\multicolumn{2}{c|}{\textbf{Fine-tuned}}&\multicolumn{3}{c|}{\textbf{Re-trained}}\\
\hline
\textbf{Model Accuracy}&\textbf{U-set}&\textbf{R-set}&\textbf{U-set}&\textbf{R-set}&\textbf{U-set}&\textbf{R-set}&\textbf{U-set}&\textbf{R-set}&\textbf{U-set}&\textbf{R-set}&\textbf{Speedup}
\\
\hline
\textbf{Delete class 0+9 from [0-9]}&
91.75\% & 83.23\% & 01.20\% & 12.50\% & 00.00\% & 19.38\% & 00.00\% & 87.22\% & 00.00\% & 86.38\% & ×10.33
\\
\hline
\textbf{Delete class 0+5+9 from [0-9]}&
85.57\% & 84.66\% & 03.82\% & 14.29\% & 00.00\% & 33.33\% & 00.00\% & 87.10\% & 00.00\% & 87.23\% & ×9.66
\\
\hline
\bottomrule
\end{tabular}
\begin{tabular}{|c|c|c|c|c|c|c|c|c|c|c|c|}
\hline
VGG11 CIFAR10 &\multicolumn{2}{c|}{\textbf{Raw model}}&\multicolumn{2}{c|}{\textbf{First class pruned}}&\multicolumn{2}{c|}{\textbf{Last class pruned}}&\multicolumn{2}{c|}{\textbf{Fine-tuned}}&\multicolumn{3}{c|}{\textbf{Re-trained}}\\
\hline
\textbf{Model Accuracy}&\textbf{U-set}&\textbf{R-set}&\textbf{U-set}&\textbf{R-set}&\textbf{U-set}&\textbf{R-set}&\textbf{U-set}&\textbf{R-set}&\textbf{U-set}&\textbf{R-set}&\textbf{Speedup}
\\
\hline
\textbf{Delete class 0+5+9 from [0-9]}&
83.60\% & 85.70\% & 00.63\% & 17.77\% & 00.00\% & 20.61\% & 00.00\% & 88.40\% & 00.00\% & 88.34\% & ×10.77
\\
\hline
\bottomrule
\end{tabular}
\begin{tabular}{|c|c|c|c|c|c|c|c|c|c|c|c|}
\hline
VGG19 CIFAR10 &\multicolumn{2}{c|}{\textbf{Raw model}}&\multicolumn{2}{c|}{\textbf{First class pruned}}&\multicolumn{2}{c|}{\textbf{Last class pruned}}&\multicolumn{2}{c|}{\textbf{Fine-tuned}}&\multicolumn{3}{c|}{\textbf{Re-trained}}\\
\hline
\textbf{Model Accuracy}&\textbf{U-set}&\textbf{R-set}&\textbf{U-set}&\textbf{R-set}&\textbf{U-set}&\textbf{R-set}&\textbf{U-set}&\textbf{R-set}&\textbf{U-set}&\textbf{R-set}&\textbf{Speedup}
\\
\hline
\textbf{Delete class 0+5+9 from [0-9]}&
84.87\% & 83.71\% & 05.60\% & 24.13\% & 00.00\% & 34.17\% & 00.00\% & 89.44\% & 00.00\% & 89.40\% & ×9.29
\\
\hline
\bottomrule
\end{tabular}
\caption{Extensive experiments: delete multiple categories of CIFAR10 simultaneously from pre-trained models.}
\label{tab:multi_full}
\end{table*}

\begin{algorithm}[ht] 
\caption{Framework of federated unlearning.} 
\label{alg:framework} 
\begin{algorithmic}[1]
\REQUIRE
the collection of participant FL clients, $\bm{n}$;\\
the FL model that has been pre-trained, $\bm{w}$;\\
the set of classes to be forgotten, $\bm{U}'$;
\ENSURE
the FL model with target classes unlearned, $\bm{w}'$;
\IF{there is a request to remove $\bm{U}'$ from $\bm{w}$} 
\STATE the federated server notifies all clients in $\bm{n}$;\\
\FOR{each client $i$ in $\bm{n}$}
\STATE download unlearning program from the server;\\
\STATE calculate \texttt{local\_proc}($\bm{w}$, private images);\\
/* defined in Section~\ref{subsec:local_proc} */
\STATE upload calculated representation to the server;
\ENDFOR
\IF{the server got uploads from all clients in $\bm{n}$} 
\STATE pruned $\hat{\bm{w}}\leftarrow$  \texttt{server\_proc}($\bm{w}$, $u'\in\bm{U}'$, aggregation);\\
/* defined in Section~\ref{subsec:server_proc} */
\ENDIF
\STATE $\bm{w}'\leftarrow$ \texttt{finetuning\_proc}($\hat{\bm{w}}$, target accuracy);\\
\RETURN $\bm{w}'$;
\ENDIF
\end{algorithmic}
\end{algorithm}

\end{document}